\documentclass[10pt,twocolumn,letterpaper]{article}

\usepackage[pagenumbers]{iccv} % To force page numbers, e.g. for an arXiv version

\usepackage{multirow}
\usepackage{amssymb}
\usepackage{amsmath}

\DeclareMathOperator*{\argmin}{arg\,min}

\definecolor{iccvblue}{rgb}{0.21,0.49,0.74}
\usepackage[pagebackref,breaklinks,colorlinks,allcolors=iccvblue]{hyperref}

\def\paperID{2440} % *** Enter the Paper ID here
\def\confName{ICCV}
\def\confYear{2025}

\title{C4D: 4D Made from 3D through Dual Correspondences}

\author{
Shizun Wang\textsuperscript{1} \quad 
Zhenxiang Jiang\textsuperscript{1} \quad  
Xingyi Yang\textsuperscript{2} \quad  
Xinchao Wang\textsuperscript{1}\thanks{Corresponding author.} \\
\vspace{-3mm} \\
\textsuperscript{1}National University of Singapore \quad 
\textsuperscript{2}The Hong Kong Polytechnic University \\
{\tt\small \{shizun.wang, zhenxiang.jiang\}@nus.u.edu,  xingyi.yang@polyu.edu.hk,  xinchao@nus.edu.sg}
}

\begin{document}
% \maketitle

% make teaser
\twocolumn[{
 \renewcommand\twocolumn[1][]{#1}% 
 \maketitle
 \begin{center}
 \vspace{-7mm}
  \captionsetup{type=figure}
  \includegraphics[width=\textwidth]{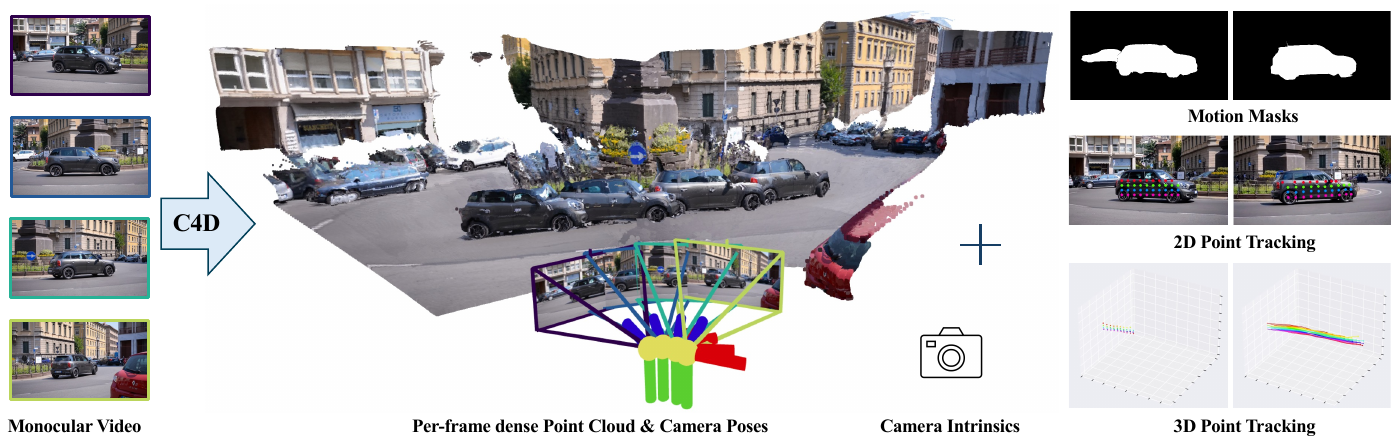}
  \captionof{figure}{Given a monocular video that contains both camera movement and object movement, \textbf{C4D} can recover the dynamic scene in 4D, including per-frame dense point cloud, camera poses and intrinsic parameters. Video depth, motion masks, and point tracking in both 2D and 3D space are also available in the outputs.}
  \label{fig:teaser}
 \end{center}
}] 

 \renewcommand{\thefootnote}{\fnsymbol{footnote}}
% % \footnotetext[1]{Equal contribution.}
 \footnotetext[1]{Corresponding author.}

\begin{abstract}

Recovering 4D from monocular video, which jointly estimates dynamic geometry and camera poses, is an inevitably challenging problem. While recent pointmap-based 3D reconstruction methods (e.g., DUSt3R) have made great progress in reconstructing static scenes, directly applying them to dynamic scenes leads to inaccurate results. This discrepancy arises because moving objects violate multi-view geometric constraints, disrupting the reconstruction.
To address this, we introduce \underline{\textbf{C4D}}, a framework that leverages temporal \underline{\textbf{C}}orrespondences to extend existing 3D reconstruction formulation to \underline{\textbf{4D}}. Specifically, apart from predicting pointmaps, C4D captures two types of \textit{correspondences}: \textit{short-term} optical flow and \textit{long-term} point tracking. We train a dynamic-aware point tracker that provides additional mobility information, facilitating the estimation of motion masks to separate moving elements from the static background, thus offering more reliable guidance for dynamic scenes.
Furthermore, we introduce a set of dynamic scene optimization objectives to recover per-frame 3D geometry and camera parameters. Simultaneously, the correspondences lift 2D trajectories into smooth 3D trajectories, enabling fully integrated 4D reconstruction.
Experiments show that our framework achieves complete 4D recovery and demonstrates strong performance across multiple downstream tasks, including depth estimation, camera pose estimation, and point tracking.
Project Page: \href{https://littlepure2333.github.io/C4D}{https://littlepure2333.github.io/C4D}

\end{abstract}    
% \vspace{-3mm}
\section{Introduction}
\label{sec:intro}

Recovering complete 4D representations from monocular videos, which involves estimating dynamic scene geometry, camera poses, and 3D point tracking, is a highly challenging task. 
While extending 3D reconstruction methods over the time dimension might seem straightforward, achieving accurate and smooth time-varying geometries and consistent camera pose trajectories is far from simple.

Recent paradigm shifts in 3D reconstruction, such as DUSt3R~\cite{wang2024dust3r}, have shown significant success in reconstructing static scenes from unordered images. It directly predicts dense 3D pointmaps from images and makes many 3D downstream tasks, like recovering camera parameters and global 3D reconstruction, become easy by just applying global alignment optimization on 3D pointmaps.

However, when applied to dynamic scenes, these formulations often produce substantial inaccuracies. This is because their reliance on multi-view geometric constraints breaks down as moving objects violate the assumptions of global alignment. As a result, they struggle to achieve accurate 4D reconstructions in dynamic scenes.

Our key insight is that the interplay between temporal correspondences and 3D reconstruction naturally leads to 4D. By capturing 2D correspondences over time, we can effectively separate moving regions from static ones. By calibrating the camera in the static region only, we improve the quality of the 3D reconstruction. In turn, the improved 3D model helps connect these correspondences, creating a consistent 4D representation that integrates temporal details into the 3D structure.

This motivation drives our framework, \textit{\textbf{C4D}}, a framework designed to upgrade the current 3D reconstruction formulation by using temporal \textit{\underline{\textbf{C}}orrespondences} to achieve \textit{\underline{\textbf{4D}}} reconstruction. Apart from 3D pointmap prediction, C4D captures \textit{short-term} optical flow and \textit{long-term} point tracking. These temporal correspondences are essential: 
they generate motion masks that guide the 3D reconstruction process, while also contributing to optimizing the smoothness of the 4D representation.

To achieve this, we introduce the Dynamic-aware Point Tracker (DynPT), which not only tracks points but also predicts whether they are moving in the world coordinates. Using this information, we create a correspondence-guided strategy that combines static points and optical flow to generate motion masks. These motion masks guide the 3D reconstruction by focusing on static regions, enabling more accurate estimation of camera parameters from the point maps and further enhancing geometric consistency.

To further improve the 4D reconstruction, we introduce a set of correspondence-aided optimization techniques. These include ensuring the camera movements are consistent, keeping the camera path smooth, and maintaining smooth trajectories for the 3D points. Together, these improvements result in a refined and stable 4D reconstruction that is both accurate and smooth over time. Extensive experiments show that C4D delivers strong performance in dynamic scene reconstruction. When applied to various downstream tasks, such as depth estimation, camera pose estimation, and point tracking, C4D performs competitively, even compared to specialized methods.

In summary, our key contributions are as follows:
\begin{itemize}
    \item We introduce C4D, a framework that upgrades the current 3D reconstruction formulation to 4D reconstruction by incorporating two temporal correspondences.
    \item We propose a dynamic-aware point tracker (DynPT) that not only tracks points but also predicts whether a point is dynamic in world coordinates.
    \item We present a motion mask prediction mechanism guided by optical flow and our DynPT.
    \item We introduce correspondence-aided optimization techniques to improve the consistency and smoothness of 4D reconstruction.
    \item We conduct experiments on depth estimation, camera pose estimation, and point tracking, demonstrating that C4D achieves strong performance, even compared to specialized methods.
\end{itemize}

\begin{figure*}[t]
\centering
\includegraphics[width=\textwidth]{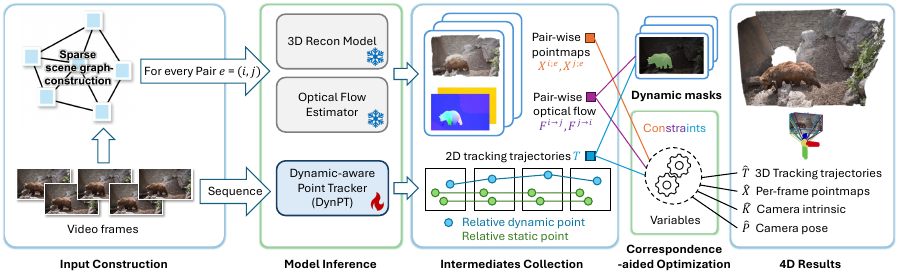}
\caption{\textbf{Overview of C4D.} 
C4D takes monocular video as input and jointly predicts dense 3D pointmaps (Sec.~\ref{sec:3d}) and temporal correspondences (Sec.~\ref{sec:correspondence}), including short-term optical flow and long-term point tracking (Sec.~\ref{sec:track}). These correspondences are utilized to predict motion masks (Sec.~\ref{sec:motion_mask}) and participate in the optimization process (Sec.~\ref{sec:optimization}) with 3D pointmaps to obtain 4D outputs.}
\label{fig:method}
\vspace{-3mm}
\end{figure*}

\section{Related Work}

\subsection{Temporal Correspondences}

\textbf{Optical flow} represents dense pixel-level motion displacement between consecutive frames, capturing short-term dense correspondences. 
Modern deep learning methods have transformed optical flow estimation, leveraging large datasets~\cite{mayer2016large, sintel}, CNNs~\cite{dosovitskiy2015flownet, sun2018pwc}, ViTs~\cite{xu2022gmflow}, and iterative refinement~\cite{teed2020raft, wang2025searaft}, resulting in significant improvements in accuracy and robustness. We leverage the motion information contained in optical flow to generate motion masks in this work.
%-------------------------------------------------------------------------
\textbf{Point tracking} aims to track a set of query points and predict their position and occlusion in a video~\cite{doersch2022tap}, providing long-term sparse pixel correspondences. 
Tracking Any Point (TAP) methods~\cite{harley2022particle, doersch2023tapir, karaev2023cotracker, locotrack} extract correlation maps between frames and use a neural network to predict tracking positions and occlusions, achieving strong performance on causal videos. While these methods are effective, they all lack the ability to predict the mobility of points in world coordinates, which we achieve in this work.

%-------------------------------------------------------------------------
\subsection{3D Reconstruction}
Recovering 3D structures and camera poses from image collections has been studied for decades~\cite{hartley2003multiple}.
Classic methods such as Structure-from-motion (SfM)~\cite{schonberger2016structure} and visual SLAM~\cite{davison2007monoslam, newcombe2011dtam} operate in sequential pipelines, often involving keypoint detection~\cite{bay2008speeded, lowe1999object, lowe2004distinctive, rublee2011orb}, matching~\cite{wu2013towards, sameer2009building}, triangulation, and bundle adjustment~\cite{agarwal2010bundle, triggs2000bundle}.
However, the sequential pipeline is complex and vulnerable to errors in each sub-task. To address these, DUSt3R~\cite{wang2024dust3r} introduces a significant paradigm shift by directly predicting pointmaps from image pairs, and dense 3D reconstruction can be obtained by a global alignment optimization.

%-------------------------------------------------------------------------
\subsection{4D Reconstruction}
Since the world is dynamic, 4D reconstruction naturally extends 3D reconstruction. Recent works~\cite{wang2024gflow, wang2024shape, lei2024mosca, stearns2024dynamic, liu2024modgs, chu2024dreamscene4d, kong2025efficient, kong2025generative, chen2025easi3r, han2025d, sucar2025dynamic, rogsplat} explore 4D reconstruction from monocular video. Building on either 3DGS~\cite{kerbl20233d} or pointmap~\cite{wang2024dust3r} representation, most of these methods are optimization-based and rely on off-the-shelf priors for supervision, such as depth, optical flow, and tracking trajectories. Concurrent work MonST3R~\cite{zhang2024monst3r} explores pointmap-based 4D reconstruction by fine-tuning DUSt3R on dynamic scene data, whereas we directly use pretrained pointmap-based model weights and complement them with correspondence-guided optimization for 4D reconstruction.

\section{Method}

The core idea of our method is to jointly predict dense 3D pointmaps and temporal correspondences from an input video, leveraging these correspondences to improve 4D reconstruction in dynamic scenes. These correspondences are obtained from both \textit{short-term} optical flow and \textit{long-term} point tracking. The whole pipeline is shown in Figure~\ref{fig:method}.

We begin by reviewing the 3D reconstruction formulation in Sec.~\ref{sec:3d}, which provides dense 3D pointmaps. Next, we introduce our dynamic-aware point tracker (DynPT) in Sec.~\ref{sec:track}, designed to track points while also identifying whether they are dynamic in world coordinates. In Sec.~\ref{sec:motion_mask}, we describe how DynPT is combined with optical flow to estimate reliable motion masks. Finally, Sec.~\ref{sec:optimization} details our correspondence-aided optimization, which utilizes pointmaps, optical flow, point tracks, and motion masks to refine the 4D reconstruction.

\subsection{3D Reconstruction Formulation}
\label{sec:3d}
Our method complements the recent feed-forward 3D reconstruction paradigm, DUSt3R~\cite{wang2024dust3r}, and can be applied to any DUSt3R-based model weights~\cite{leroy2024mast3r, zhang2024monst3r}. Given a video with $T$ frames $\{I^1, I^2,...,I^T\}$,
a scene graph $\mathcal{G}$ is constructed, where an edge represents a pair of images $e = (I^n,I^m) \in \mathcal{G}$.
Then DUSt3R operates in two steps:

(1) A ViT-based network $\Phi$ that takes a pair of two images $I^n,I^m \in \mathbb{R}^{W\times H \times 3}$ as input and directly outputs two dense pointmaps $X^n,X^m \in \mathbb{R}^{W\times H \times 3}$ with associated confidence maps $C^n,C^m \in \mathbb{R}^{W\times H}$.
\begin{equation}
    X^n, C^n, X^m, C^m = \Phi(I^n, I^m)
\end{equation}

(2) Since these pointmaps are represented in the local coordinate of each pair, DUSt3R employs global optimization to all pairs of pointmaps, to recover \textit{global aligned} pointmaps $\{\mathcal{X}^t \in \mathbb{R}^{W\times H \times 3}\}$ for all frames $t = 1,...,T$.
\begin{equation}
     \mathcal{L}_{\textrm{GA}} (\mathcal{X}, P, \sigma) = 
     \sum_{e\in \mathcal{G}} \sum_{t\in e} 
     \mathbf{C}^{t;e} ||\mathcal{X}^t - \sigma_{e} P_e X^{t; e} ||
\end{equation}

Where $P_e \in \mathbb{R}^{3\times4}$ and $\sigma_e > 0$ are pairwise pose and scaling. 
To reduce computational cost, we use a sparse scene graph based on a strided sliding window, as in~\cite{wang2024dust3r, duisterhof2024mast3r, zhang2024monst3r}, where only pairs within a local temporal window are used for optimization.

While this 3D formulation performs well on static scenes, its performance drops in dynamic scenes, as discussed in Sec.~\ref{sec:improvement}. This is primarily due to moving objects violating multi-view geometric constraints, which motivates us to extend the current 3D formulation to a 4D one.

\subsection{Capturing Dual Correspondences}
\label{sec:correspondence}
We capture two correspondences to help 4D recovery: \textit{long-term} point tracking and \textit{short-term} optical flow.
\subsubsection{Dynamic-aware Point Tracker}
\label{sec:track}
Current 2D point tracking methods like Tracking Any Point (TAP)~\cite{doersch2022tap, doersch2023tapir, karaev2023cotracker, karaev2024cotracker3} can robustly track query points in videos. 
However, they cannot distinguish whether the movement of the tracking point is caused by camera movement or object movement.
To segment moving objects in the \emph{world coordinate} system, we enhance these trackers by enabling them to predict the mobility of tracking points. We introduce the \textbf{Dynamic-aware Point Tracker (DynPT)}, which differentiates between motion caused by the camera and true object dynamics. This helps us identify and segment moving objects even when both the camera and the objects are in motion.

\begin{figure}[t]
\includegraphics[width=\linewidth]{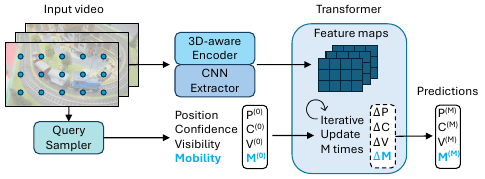}
\caption{\textbf{Architecture of Dynamic-aware Point Tracker (DynPT).} For given video input and sampled initial query points, DynPT uses Transformer to iteratively update the tracks with features obtained from both 3D-aware ViT encoder and CNN.}
\label{fig:tracker}
\vspace{-3mm}
\end{figure}

\textbf{Tracker Architecture  }
We adopt a similar design of CoTracker~\cite{karaev2023cotracker, karaev2024cotracker3} to design our DynPT, which is illustrated in Figure~\ref{fig:tracker}. Original CoTracker only uses one CNN~\cite{resnet} to extract features. While to better capture the spatial dynamic relationships, we additionally employ a 3D-aware ViT encoder, which comes from DUSt3R's encoder, to enhance the 3D spatial information~\cite{weinzaepfel2023croco}. And different from all other TAP methods, DynPT directly predicts one more attribute, mobility, along with other attributes of tracks. 

Specifically, for an input video of length $T$, DynPT first extracts each frame's multi-scale features from a 3D-aware encoder and CNN, which are used to construct 4D correlation features $Corr$ to provide richer information for tracking~\cite{locotrack}. Given a query point $P_0 \in \mathbb{R}^2$ at the first frame, we initialize the track positions $P_t$ with the same position of $P_0$ for all remaining times $t=1,...,T$, and initialize the confidence $C_t$, visibility $V_t$ and mobility $M_t$ with zeros for all times. Then we iteratively update these attributes with a transformer for $M$ times. At each iteration, the transformer takes a grid of input tokens spanning time $T$: $G^i_t = ( \eta^i_{t-1 \rightarrow t}, \eta^i_{t \rightarrow t+1}, C^i_t, V^i_t, M^i_t, Corr^i_t)$ for every query point $i=1,...,N$, where $\eta^i_{t \rightarrow t+1} = \eta (P_{t+1} - P_t)$ represents Fourier Encoded embedding of per-frame displacements. Inside the Transformer, the attention operation is applied across time and track dimensions.

\textbf{Training and Inference}
We train DynPT on Kubric~\cite{greff2022kubric}, a synthetic dataset from which ground-truth mobility labels can be obtained.  We use Huber loss to supervise position. And we employ cross-entropy loss to supervise confidence, visibility and mobility. When performing inference on a video, DynPT predicts tracks in a sliding window manner. 
More details about the DynPT can be found in the supplementary materials.

\begin{figure}[t]
\includegraphics[width=\linewidth]{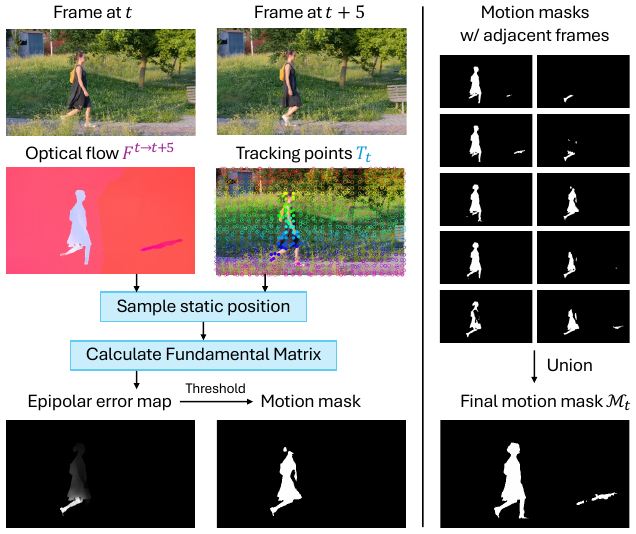}
\caption{\textbf{Correspondence-guided motion mask prediction.} The solid circle indicates predicted dynamic points, the hollow circle indicates predicted static points. Adjacent frames are from constructed image pairs containing the current frame.}
\label{fig:dynamic_mask}
\vspace{-3mm}
\end{figure}

\subsubsection{Correspondence-Guided Motion Mask Estimation}
\label{sec:motion_mask}
The most important part of 4D reconstruction in dynamic scenes is to separate dynamic areas from static areas in world coordinates.
To achieve this, we utilize two temporal correspondences: short-term optical flow $F_{est}$ estimated by off-the-shelf models~\cite{teed2020raft, wang2025searaft, xu2022gmflow}, and long-term point tracking trajectory $T$ predicted by DynPT. Figure~\ref{fig:dynamic_mask} shows this strategy of correspondence-guided motion mask prediction.

Since DynPT provides mobility predictions of tracks, at time $t$, we can retrieve the positions of static points $\{P_t^j\}$ where $M_t^j=0$. And given an optical flow $F^{t \rightarrow t'}$ from time $t$ to adjacent time $t'$, we can sample the pixel correspondences of these static points $\{(P_t^j,P_{t'}^j)\}$. With these correspondences, we then estimate the fundamental matrix $\mathcal{F}$ between the two frames via the Least Median of Squares (LMedS) method~\cite{rousseeuw1984least}, which does not require known camera parameters and is robust to outliers. 
Since the fundamental matrix is estimated solely based on static points, it reflects only the underlying camera motion, unaffected by dynamic objects in the scene.
So using this $\mathcal{F}$ to calculate the epipolar error map on all correspondences in $F^{t \rightarrow t'}$, the area with large error indicates there violates the epipolar constraints, that is, dynamic area. In practice, we compute the error map using the Sampson error~\cite{hartley2003multiple}, which provides a more robust approximation of the epipolar error by accounting for scale and orientation. Then a threshold is applied to obtain the motion mask.

While considering a longer temporal range, calculating the motion mask based on only two frames is not sufficient. For example, a person's standing foot may remain still for several frames before lifting off to step, as shown in Figure~\ref{fig:dynamic_mask}. To address this, we calculate the motion mask of the current frame using adjacent frames from the constructed image pairs that include the current frame $t$, then take the union of these masks to produce the final motion mask $\mathcal{M}_t$.

\subsection{Correspondence-aided Optimization for 4D}
\label{sec:optimization}

Based on the \textbf{Global Alignment (GA)} objective described in Sec~\ref{sec:3d}, we introduce additional optimization objectives to improve the accuracy and smoothness in dynamic scenes: \textit{camera movement alignment}, \textit{camera trajectory smoothness}, and \textit{point trajectory smoothness}.
The optimizable variables are per-frame depthmap $D^t$, camera intrinsic $K^t$ and camera pose $P^t= [ R^t | T^t ]$. Then we re-parameterize the global pointmaps $\mathcal{X}^t$ as $\mathcal{X}_{i,j}^{t} := {P^t}^{-1} h ({K^{t}}^{-1} [ i D_{i,j}^{t}; j D_{i,j}^{t}; D_{i,j}^{t} ] )$, where $(i,j)$ is the pixel coordinate and $h(\cdot)$ is the homogeneous mapping. So that, optimizing $\mathcal{X}^t$ is equivalent to optimizing $P^t, K^t, D^t$.

Since global alignment tends to align moving objects to the same position, it can negatively impact camera pose estimation. To address this, and leveraging the fact that optical flow provides a prior on camera motion, we introduce the \textbf{Camera Movement Alignment (CMA)} objective~\cite{zhang2024monst3r, wang2024gflow, sun2024splatter, zhang2021consistent, kappel2024d}. CMA encourages the estimated ego motion to be consistent with optical flow in static regions.
Specifically, for two frames $I^t$ and $I^{t'}$, we compute the ego-motion field $F_{ego}^{t \rightarrow t'}$ as the 2D displacement of $\mathcal{X}^t$ by moving camera from $t$ to $t'$.
Then we can encourage this field to be close to the optical flow field $\textbf{F}^{t\rightarrow t'}$, in the static regions $\mathcal{S}_t = \sim \mathcal{M}_t$, which is the opposite region of the motion mask $\mathcal{M}_t$ computed in Sec~\ref{sec:motion_mask}:
\begin{equation}
\mathcal{L}_{\text{CMA}}(\mathcal{X}) = \sum_{e \in \mathcal{G}} \sum_{(t,t') \in e} \left\| \mathcal{S} \cdot \left( \mathbf{F}_{\text{ego}}^{t\rightarrow t'} - \mathbf{F}^{t\rightarrow t'} \right) \right\|_1
\end{equation}

The \textbf{Camera Trajectory Smoothness (CTS)} objective is commonly used in visual odometry~\cite{mur2015orb, steinbrucker2011real, zhang2024monst3r} to enforce smooth camera motion by penalizing abrupt changes in camera rotation and translation between consecutive frames:
\begin{equation}
\mathcal{L}_{\text{CTS}}(\mathbf{X})  = \sum_{t=0}^{N}  \left\| {\mathbf{R}^t}^\top \mathbf{R}^{t+1} - \mathbf{I} \right\|_\text{F} + \left\|  (\mathbf{T}^{t+1} - \mathbf{T}^t) \right\|_2 
\end{equation}

where $\| \cdot \|_\text{F}$ denotes the Frobenius norm, and $\mathbf{I}$ is the identity matrix.

Lastly, we propose the \textbf{Point Trajectory Smoothness (PTS)} objective to smooth world coordinate pointmaps over time. Within a local temporal window, we first select 2D tracking trajectories $T$ that remain visible throughout the window and lift them to 3D trajectories. We then smooth these 3D trajectories using a 1D convolution with adaptive weights, where weights are reduced for outlier points based on their temporal deviations. For each frame within the window, we treat the smoothed points as control points and apply a linear blend of control point displacements to transform all other points, weighting each control point's influence based on proximity, resulting in dense smoothed pointmaps $\widetilde{\mathcal{X}}^t$. (More details in supplementary.)

We then minimize per-frame distance between global pointmaps and smoothed pointmaps using L1 loss:
\begin{equation}
    \mathcal{L}_{\text{PTS}}(\mathcal{X}) = \sum_{t=0}^{N} \left(||\mathcal{X}^t - \widetilde{\mathcal{X}}^t ||_1 \right)
\end{equation}

The complete optimization objective for recovering the 4D scene is:
\begin{equation}
\begin{aligned}
    \hat{\mathcal{X}}
    = &\argmin_{\mathcal{X}, P, \sigma} 
     \big[w_{\text{GA}} \mathcal{L}_\textrm{GA}(\mathcal{X}, \sigma, P)
    + w_{\text{CMA}} \mathcal{L}_\textrm{CMA}(\mathcal{X}) \\
    &+ w_{\text{CTS}} \mathcal{L}_\textrm{CTS}(\mathcal{X})
    + w_{\text{PTS}} \mathcal{L}_\textrm{PTS}(\mathcal{X}) \big]
\end{aligned}
\end{equation}
\label{eq:global_optimization}
where $w_{\text{GA}}, w_{\text{CMA}}, w_{\text{CTS}}, w_{\text{PTS}}$ are the loss weights. 
The completed outputs of C4D contain world-coordinate pointmaps $\hat{\mathcal{X}}$, depthmaps $\hat{D}$, camera poses $\hat{P}$, camera intrinsics $\hat{K}$, motion masks $\mathcal{M}$, 2D tracking trajectories $T$, and lifted 3D tracking trajectories $\hat{\mathbf{T}}$.

\section{Experiments}
% \subsection{Implementation Details}
We evaluate C4D on multiple downstream tasks, comparing it with specialized methods (Sec.~\ref{sec:compare}), and 3D formulations (Sec.~\ref{sec:improvement}). The ablation study in Sec.~\ref{sec:ablation} justifies our design choices, and implementation details are provided in supplementary materials.

\begin{table*}[t]
\centering
\footnotesize
\renewcommand{\arraystretch}{1.3}
\renewcommand{\tabcolsep}{2.5pt}

\begin{tabular}{@{}clccc|ccc|ccc@{}}
\toprule
\multirow{2}{*}{\begin{minipage}{2.0cm}\centering 3D Model Weights\end{minipage}}&\multirow{2}{*}{\begin{minipage}{2.0cm} Optimization Formulation\end{minipage}} & \multicolumn{3}{c}{Sintel} & \multicolumn{3}{c}{TUM-dynamics} & \multicolumn{3}{c}{ScanNet (static)} \\ 
\cmidrule(lr){3-5} \cmidrule(lr){6-8} \cmidrule(lr){9-11}
& & {ATE $\downarrow$} & {RPE trans $\downarrow$} & {RPE rot $\downarrow$} & {ATE $\downarrow$} & {RPE trans $\downarrow$} & {RPE rot $\downarrow$} & {ATE $\downarrow$} & {RPE trans $\downarrow$} & {RPE rot $\downarrow$} \\ 
\midrule
\multirow{2}{*}{DUSt3R} 
& Global Alignment & 0.416 & 0.216 & 18.038 & 0.127 & 0.058 & 2.033 & \textbf{0.060} & 0.024 & 0.751 \\ 
& \textbf{C4D} & \textbf{0.334} & \textbf{0.154} & \textbf{0.948} & \textbf{0.093}  & \textbf{0.018}  & \textbf{0.906}  & 0.064  & \textbf{0.018}  & \textbf{0.570}  \\ 
\midrule
\multirow{2}{*}{MASt3R} 
& Global Alignment & \textbf{0.437} & 0.329 & 12.760 & 0.084 & 0.052 & 1.245 & 0.073 & 0.027 & 0.706 \\ 
& \textbf{C4D} & 0.448 & \textbf{0.199} & \textbf{1.730} & \textbf{0.048}  & \textbf{0.012}  & \textbf{0.671}  & \textbf{0.067}  & \textbf{0.018}  &  \textbf{0.467} \\ 
\midrule
\multirow{2}{*}{MonSt3R} 
& Global Alignment & 0.158 & 0.099 & 1.924 & 0.099 & 0.041 & 1.912 & 0.075 & 0.026 & 0.707 \\ 
& \textbf{C4D (Ours)} & \textbf{0.103} & \textbf{0.040} & \textbf{0.705} & \textbf{0.071}  & \textbf{0.019} & \textbf{0.897}  & \textbf{0.061}  & \textbf{0.017}  & \textbf{0.538}  \\ 
\bottomrule
\end{tabular}
\caption{\textbf{Camera pose estimation results across 3D/4D formulations.} Evaluation on the Sintel, TUM-dynamic, and ScanNet datasets. The best results are highlighted in \textbf{bold}. Our 4D formulation, C4D, consistently improves the performance based on 3D models.}
\label{tab:camera_pose_3d}

\end{table*}

\begin{table*}[t]
\centering
\footnotesize
\renewcommand{\arraystretch}{1.2}
\renewcommand{\tabcolsep}{1.5pt}

\begin{tabular}{@{}clccc|ccc|ccc@{}}
\toprule
\multirow{2}{*}{\begin{minipage}{2.0cm}\centering 3D Model Weights\end{minipage}}&\multirow{2}{*}{\begin{minipage}{2.0cm} Optimization Formulation\end{minipage}} & \multicolumn{3}{c}{Sintel} & \multicolumn{3}{c}{Bonn} & \multicolumn{3}{c}{KITTI} \\ 
\cmidrule(lr){3-5} \cmidrule(lr){6-8} \cmidrule(lr){9-11}
& & {Abs Rel $\downarrow$} & RMSE $\downarrow$ & {$\delta$\textless $1.25\uparrow$} & {Abs Rel $\downarrow$} & RMSE $\downarrow$ & {$\delta$\textless $1.25\uparrow$} & {Abs Rel $\downarrow$} & RMSE $\downarrow$ & {$\delta$\textless $1.25\uparrow$} \\ 
\midrule
\multirow{2}{*}{DUSt3R} 
& Global Alignment & 0.502 & 5.141 & 54.9 & 0.149  & 0.422 & 84.4 & 0.129  & 5.162 & 84.2 \\ 
& \textbf{C4D (Ours)} & \textbf{0.478} & \textbf{5.052} & \textbf{57.9} & \textbf{0.143}  & \textbf{0.411}  & \textbf{84.7}  & \textbf{0.126}  & \textbf{5.140}  &  \textbf{85.0}  \\ 
\midrule
\multirow{2}{*}{MASt3R} 
& Global Alignment & \textbf{0.370}  & \textbf{4.669} & 57.8 & 0.174 & 0.503 & 78.4 & 0.092 & 4.000  & \textbf{89.8}  \\ 
& \textbf{C4D (Ours)} & 0.379 & 4.756 & \textbf{58.3} & \textbf{0.168} &  \textbf{0.485} & \textbf{78.6}  & \textbf{0.092}  &  \textbf{4.000} & 89.7  \\ 
\midrule
\multirow{2}{*}{MonSt3R} 
& Global Alignment & 0.335  & 4.467 & 57.5 & 0.065 & 0.254 & 96.2 & 0.090 & 4.128 & 90.6  \\ 
& \textbf{C4D (Ours)} & \textbf{0.327} & \textbf{4.465} & \textbf{60.7} & \textbf{0.061} & \textbf{0.249} & \textbf{96.5} & \textbf{0.089} & \textbf{4.128} & \textbf{90.6} \\ 
\bottomrule
\end{tabular}
\caption{\textbf{Video depth estimation results across 3D/4D formulations.} We evaluate scale-and-shift-invariant depth on Sintel, Bonn, and KITTI. The best results are highlighted in \textbf{bold}. Our 4D fomulation, C4D, consistently improve the performance based on 3D models.}
\label{tab:video_depth_3d}
% \vspace{-1em}

\end{table*}

\begin{table*}
\centering
\footnotesize
\renewcommand{\arraystretch}{1.3}
\renewcommand{\tabcolsep}{2.5pt}

\begin{tabular}{@{}clccc|ccc|ccc@{}}
\toprule
\multirow{2}{*}{{Category}}& \multirow{2}{*}{{Method}} & \multicolumn{3}{c}{Sintel} & \multicolumn{3}{c}{TUM-dynamics} & \multicolumn{3}{c}{ScanNet (static)} \\ 
\cmidrule(lr){3-5} \cmidrule(lr){6-8} \cmidrule(lr){9-11}
  &   & {ATE $\downarrow$} & {RPE trans $\downarrow$} & {RPE rot $\downarrow$} & {ATE $\downarrow$} & {RPE trans $\downarrow$} & {RPE rot $\downarrow$} & {ATE $\downarrow$} & {RPE trans $\downarrow$} & {RPE rot $\downarrow$} \\ 
\midrule
\multirow{4}{*}{{Pose only}} 
& DROID-SLAM$^\dagger$ & 0.175 & 0.084 & 1.912 & - & - & - & - & - & - \\ 
& DPVO$^\dagger$ & 0.115 & 0.072 & 1.975 & - & - & - & - & - & - \\ 
& ParticleSfM & 0.129 & \textbf{0.031} & \textbf{0.535} & - & - & - & 0.136 & 0.023 & 0.836 \\ 
& LEAP-VO$^\dagger$ & \textbf{0.089} & 0.066 & 1.250 & {\underline{0.068}} & \textbf{0.008} & 1.686 & {\underline{0.070}} & {0.018} & {\textbf{0.535}} \\ 
\midrule
\multirow{3}{*}{{Joint depth \ \& pose}} 
& Robust-CVD & 0.360 & 0.154 & 3.443 & 0.153 & 0.026 & 3.528 & 0.227 & 0.064 & 7.374 \\ 
& CasualSAM & 0.141 & \underline{0.035} & \underline{0.615} & 0.071 & \underline{0.010} & 1.712 & 0.158 & 0.034 & 1.618 \\ 
& MonST3R & {0.109} & {0.043} & {0.737} & {0.104} & {0.223} & \underline{1.037} & \underline{0.068} & \underline{0.017} & {0.545} \\ 
& \textbf{C4D-M (Ours)} & {\underline{0.103}} & {0.040} & {0.705} & \underline{0.071} & 0.019 & \textbf{0.897} & {\textbf{0.061}} & {\textbf{0.017}} & {\underline{0.538}} \\ 
\bottomrule
\end{tabular}
\caption{\textbf{Camera pose evaluation} on Sintel, TUM-dynamic, and ScanNet. The best and second best results are highlighted in \textbf{bold} and \underline{underlined}, respectively. $^\dagger$ means the method requires ground truth camera intrinsics as input. ``C4D-M'' denotes C4D with MonST3R's model weights.}
\label{tab:camera_pose}
\end{table*}

% \caption{\textbf{Camera pose evaluation} on Sintel, TUM-dynamic, and ScanNet. The best and second best results are highlighted in \textbf{bold} and \underline{underlined}, respectively. $^\dagger$ means the method requires ground truth camera intrinsics as input. ``C4D-M'' denotes C4D with MonST3R's model weights. Note that C4D requires only monocular video as input to estimate camera intrinsics and poses, while achieving comparable or even superior results to methods that rely on ground-truth camera intrinsics.}

\begin{table*}
\centering
\footnotesize
\renewcommand{\arraystretch}{1.2}
\renewcommand{\tabcolsep}{1.5pt}

% \resizebox{1\textwidth}{!}{
\begin{tabular}{@{}cclcc|cc|cc@{}}
\toprule
\multirow{2}{*}{{Alignment}} & \multirow{2}{*}{{Category}} & \multirow{2}{*}{{Method}} & \multicolumn{2}{c}{Sintel} & \multicolumn{2}{c}{Bonn} & \multicolumn{2}{c}{KITTI} \\ 
\cmidrule(lr){4-5} \cmidrule(lr){6-7} \cmidrule(lr){8-9}
 &  &  & {Abs Rel $\downarrow$} & {$\delta$\textless $1.25\uparrow$} & {Abs Rel $\downarrow$} & {$\delta$\textless $1.25\uparrow$} & {Abs Rel $\downarrow$} & {$\delta$ \textless $1.25\uparrow$} \\ 
\midrule
 \multirow{9}{*}{\begin{minipage}{2.0cm}\centering Per-sequence scale \& shift\end{minipage}} &\multirow{2}{*}{Single-frame depth} & Marigold & 0.532 & 51.5 & 0.091 & 93.1 & 0.149 & 79.6 \\ 
 && DepthAnything-V2 & 0.367 & 55.4 & 0.106 & 92.1 & 0.140 & 80.4 \\ 

\cmidrule{2-9}

 &\multirow{3}{*}{Video depth} & NVDS & 0.408 & 48.3 & 0.167 & 76.6 & 0.253 & 58.8 \\ 
 && ChronoDepth & 0.687 & 48.6 & 0.100 & 91.1 & 0.167 & 75.9 \\ 
 && DepthCrafter & \textbf{0.292} & \textbf{69.7} & \underline{0.075} & \textbf{97.1} & \underline{0.110} & \underline{88.1} \\ 

\cmidrule{2-9}

 &\multirow{4}{*}{ Joint video depth \& pose } & Robust-CVD & 0.703 & 47.8 & - & - & - & - \\ 
 && CasualSAM & 0.387 & 54.7 & 0.169 & 73.7 & 0.246 & 62.2 \\ 
& & MonST3R & 0.335 & 58.5 & 0.063 & 96.2 & 0.157 & 73.8 \\ 
& & \textbf{C4D-M (Ours)} & \underline{0.327} & \underline{60.7} & \textbf{0.061} & \underline{96.5} & \textbf{0.089} & \textbf{90.6} \\ 
\addlinespace[1.5pt]
\hline\hline
\addlinespace[1.5pt]
 \multirow{3}{*}{\begin{minipage}{2.0cm}\centering Per-sequence scale\end{minipage}} & Video depth & DepthCrafter & 0.692 & 53.5 & 0.217 & 57.6 & \underline{0.141} & \underline{81.8} \\ 

 & Joint video depth \& pose & MonST3R & \underline{0.345} & \underline{55.8} & \underline{0.065} & \underline{96.2} & 0.159 & 73.5 \\ 
 & Joint video depth \& pose & \textbf{C4D-M (Ours)} & \textbf{0.338} & \textbf{58.1} & \textbf{0.063} & \textbf{96.4} & \textbf{0.091} & \textbf{90.6} \\ 
\bottomrule
\end{tabular}
\caption{\textbf{Video depth evaluation} on Sintel, Bonn, and KITTI. 
Two types of depth range alignment are evaluated: scale \& shift and scale-only. 
``C4D-M'' denotes C4D with MonST3R's model weights. 
}
\label{tab:video_depth}
\end{table*}

% \caption{\textbf{Video depth evaluation} on Sintel, Bonn, and KITTI. 
% Two types of depth range alignment are evaluated: scale \& shift and scale-only. 
% The best and second-best results in each alignment are highlighted in \textbf{bold} and \underline{underlined}, respectively. 
% ``C4D-M'' denotes C4D with MonST3R's model weights. 
% Since C4D estimates scale-invariant metric depth, it achieves competitive or superior results compared to methods that predict relative depths.}
\begin{table*}
\centering
\footnotesize

% \small
\begin{tabular}{lccccccccc}
\toprule
\multirow{2}{*}{\textbf{Method}} & 
% \multicolumn{3}{c}{\textbf{Kinetics}} & \multicolumn{3}{c}{\textbf{DAVIS}} & \multicolumn{3}{c}{\textbf{RGB-Stacking}} \\ 
% \cmidrule(lr){2-4} \cmidrule(lr){5-7} \cmidrule(lr){8-10} 
MOVi-E & Pan. MOVi-E & MOVi-F & \multicolumn{3}{c}{\textbf{TAP-Vid DAVIS}} & \multicolumn{3}{c}{\textbf{TAP-Vid Kinetics}} \\ 
\cmidrule(lr){2-4} \cmidrule(lr){5-7} \cmidrule(lr){8-10} 
    & D-ACC~$\uparrow$ & D-ACC~$\uparrow$ & D-ACC~$\uparrow$  & AJ~$\uparrow$ & $<\delta^x_\textrm{avg}$~$\uparrow$ & OA~$\uparrow$ &  AJ~$\uparrow$ & $<\delta^x_\textrm{avg}$~$\uparrow$ & OA~$\uparrow$  \\ 
\midrule
\multicolumn{3}{l}{\textcolor{gray}{\textit{\textrm{Predict Position \& Occlusion}}}} \\
RAFT & - & - & - & 30.0  & 46.3  & 79.6  & 34.5  & 52.5  &  79.7 \\
TAP-Net & - & - & -  & 38.4  & 53.1  &  82.3 & 46.6  &  60.9 & \textbf{85.0}  \\
PIPs & - & - & - & 39.9  & 56.0  & 81.3  & 39.1  &  55.3 & 82.9  \\
MFT &  - & - & - &  47.3 & 66.8  & 77.8  & 39.6  & 60.4  & 72.7  \\
% OmniMotion & - & - & -  & 51.7  & 67.5  & 85.3  & 55.1  & 69.6  & 89.6  \\
TAPIR & - & - & - & 56.2  & 70.0  &  86.5 & \textbf{49.6}  & \underline{64.2}  & \textbf{85.0}  \\
CoTracker &  - & - & - & \textbf{61.8}  & \textbf{76.1}  & \textbf{88.3}  & \textbf{49.6}  & \textbf{64.3}  & \underline{83.3}  \\
\midrule
\multicolumn{3}{l}{\textcolor{gray}{\textit{\textrm{Predict Position \& Occlusion \& Mobility}}}}\\
DynPT (Ours) & \textbf{87.9} & \textbf{94.1} & \textbf{91.5}  & \underline{61.6}  & \underline{75.4}  & \underline{87.4}  & \underline{47.8}  & 62.6  &  82.3 \\

\bottomrule
\end{tabular}
\caption{\textbf{Point tracking evaluation results} on the TAP-Vid and Kubric (MOVi-E, Panning MOVi-E, and MOVi-F) Datasets. Apart from achieving competitive results with SOTA TAP methods, DynPT offers a unique capability: predicting the mobility of tracking points, which is crucial for determining whether a point is dynamic in world coordinates.}
\label{tab:tracking_2d}
\end{table*}

\subsection{Datasets and Metrics}
We evaluate camera pose estimation on Sintel~\cite{sintel}, TUM-dynamics~\cite{tum} and ScanNet~\cite{dai2017scannet} following~\cite{zhang2022structure, zhao2022particlesfm, chen2024leap}. Sintel is a synthetic dataset featuring challenging motion blur and large camera movements. TUM-Dynamics and ScanNet are real-world datasets for dynamic scenes and static scenes, respectively. We report the metrics of Absolute Translation Error (ATE), Relative Translation Error (RPE trans), and Relative Rotation Error (RPE rot).

For depth estimation, we evaluate on Sintel, Bonn~\cite{palazzolo2019refusion}, and KITTI~\cite{geiger2013vision}, following~\cite{hu2024depthcrafter, zhang2024monst3r}. 
Bonn~\cite{palazzolo2019refusion} and KITTI~\cite{geiger2013vision} are real-world indoor dynamic scene and outdoor datasets.
The evaluation metrics for depth estimation are Absolute Relative Error (Abs Rel), Root Mean Squared Error (RMSE), and the percentage of inlier points \(\delta < 1.25\), as used in prior works~\cite{hu2024depthcrafter, yang2024depthv2}.

For point tracking, we evaluate our method on the TAP-Vid benchmark~\cite{doersch2022tap} and Kubric~\cite{greff2022kubric}. TAP-Vid contains videos with annotations of tracking point positions and occlusion. We use the metrics of occlusion accuracy (OA), position accuracy (\(\delta_{\text{avg}}^x\)), and average Jaccard (AJ) to evaluate this benchmark, following~\cite{doersch2023tapir, karaev2023cotracker, wang2023tracking, harley2022particle}.  
Kubric is a generator that synthesizes semi-realistic multi-object falling videos with rich annotations, including the moving status of tracking points in world coordinates. To fully evaluate the diverse dynamic patterns in the real world, we use three datasets from Kubric to assess dynamic accuracy (D-ACC):  
1) MOVi-E, which introduces simple (linear) camera movement while always ``looking at" the center point in world coordinates;  
2) Panning MOVi-E, which modifies MOVi-E with panning camera movement;  
3) MOVi-F, similar to MOVi-E but adds some random motion blur.

\subsection{Comparison across 3D/4D Formulations}
\label{sec:improvement}

% \paragraph{Baselines} 
\textbf{3D Baselines} 
We choose the currently available DUSt3R-based models as our 3D baseline models: 
1) DUSt3R~\cite{wang2024dust3r}, trained on millions of image pairs in static scenes, demonstrating impressive performance and generalization across various real-world static scenarios with different camera parameters.  
2) MASt3R~\cite{leroy2024mast3r}, the follow-up work to DUSt3R, which initializes its weights from DUSt3R and is fine-tuned on the matching task, also using large-scale data from static scenes.  
3) MonSt3R~\cite{zhang2024monst3r}, which fine-tunes the decoder and head of DUSt3R on selected dynamic scene datasets.  
The global alignment is the default optimization strategy in the 3D formulation, as described in Sec.~\ref{sec:optimization}.

\textbf{Results  }
We evaluate the results of camera pose estimation and video depth estimation, as shown in Table~\ref{tab:camera_pose_3d} and Table~\ref{tab:video_depth_3d}.  
Our C4D achieves consistent performance improvements compared to 3D formulation across different 3D model weights. For camera pose estimation, C4D significantly improves performance (e.g., reducing $RPE_r$ from 18.038 to 0.948) even on the challenging Sintel dataset, demonstrating the effectiveness of our method. The results on the ScanNet dataset, which consists of static scenes, also show that our method further enhances performance in static environments. 
C4D also outperforms 3D formulations in terms of video depth accuracy. Moreover, these results highlight a comparison among 3D model weights: DUSt3R and MASt3R perform comparably overall, while MonST3R achieves better results as it is fine-tuned on dynamic scene datasets.

\subsection{Comparison with Other Methods}
\label{sec:compare}
Since C4D produces multiple outputs, we compare our method with others specifically designed for individual tasks, including camera pose estimation, video depth estimation, and point tracking.

\textbf{Evaluation on camera pose estimation  }
We compare with methods that can predict camera pose and video depth jointly: Robust-CVD~\cite{kopf2021robust}, CasualSAM~\cite{zhang2022structure}, and the concurrent work MonST3R~\cite{zhang2024monst3r}. We re-evaluated MonST3R using their publicly available codes and checkpoints for a fair comparison. For a broader evaluation, we also compare with learning-based visual odometry methods: DROID-SLAM~\cite{teed2021droid}, DPVO~\cite{teed2024deep}, ParticleSfM~\cite{zhao2022particlesfm}, and LEAP-VO~\cite{chen2024leap}. Note that DROID-SLAM, DPVO, and LEAP-VO require ground-truth camera intrinsics as input, while our C4D can estimate camera intrinsics and camera poses using only a monocular video as input. The results are presented in Table~\ref{tab:camera_pose}, showing that C4D achieves highly competitive performance even compared to specialized visual odometry methods and generalizes well on static scenes, such as the ScanNet dataset.

\textbf{Evaluation on video depth estimation  }
Table~\ref{tab:video_depth} shows the evaluation results on video depth estimation. We compare with various kinds of depth estimation methods: single-frame depth methods such as Marigold~\cite{ke2024repurposing} and DepthAnything-V2~\cite{yang2024depthv2}, and video depth methods such as NVDS~\cite{wang2023neural}, ChronoDepth~\cite{shao2024learning}, and DepthCrafter~\cite{hu2024depthcrafter}. Note that these methods predict relative depth, which leads to inconsistencies across multiple views when projecting to world coordinates~\cite{godard2019digging}. We also compare with methods that can predict video depth and camera pose jointly: Robust-CVD~\cite{kopf2021robust}, CasualSAM~\cite{zhang2022structure}, and MonST3R~\cite{zhang2024monst3r}. The evaluation is conducted using two kinds of depth range alignment: scale \& shift, and scale-only. 
C4D achieves highly competitive results in scale \& shift alignment. However, as demonstrated in~\cite{yin2021learning}, a shift in depth will affect the x, y, and z coordinates non-uniformly when recovering the 3D geometry of a scene, resulting in shape distortions. Therefore, a more important evaluation is under scale-only alignment, where C4D achieves the best performance.

\textbf{Evaluation on point tracking   }
As part of the C4D outputs, we evaluate point tracking results in Table~\ref{tab:tracking_2d} and compare them with other TAP methods: RAFT~\cite{teed2020raft}, TAP-Net~\cite{doersch2022tap}, PIPs~\cite{harley2022particle}, MFT~\cite{neoral2024mft}, TAPIR~\cite{doersch2023tapir}, and Cotracker~\cite{karaev2023cotracker}. Note that all previous TAP methods can only predict the position and occlusion of tracking points, whereas our method can additionally predict mobility, contributing to a robust motion mask prediction as described in Sec.~\ref{sec:motion_mask}. Despite this more challenging learning objective, our method still achieves comparable performance with SOTA methods and demonstrates high accuracy in predicting mobility.

\begin{table}[t]
\centering
\footnotesize
\renewcommand{\arraystretch}{1.2}
\renewcommand{\tabcolsep}{1.5pt}
\begin{tabular}{lcccccc}
\toprule
\multirow{2}{*}{Method} & \multicolumn{3}{c}{Camera pose} & \multicolumn{3}{c}{Video depth} \\
\cmidrule(lr){2-4}\cmidrule(lr){5-7}
 & ATE $\downarrow$ & RPE\_t $\downarrow$ & RPE\_r $\downarrow$ & Abs Rel $\downarrow$ & RMSE $\downarrow$ &  $\delta \textless 1.25 \uparrow$ \\
\midrule
w/o CMA    & 0.140 & 0.051 & 0.905 & 0.335 & 4.501 & 0.582 \\
w/o CTS    & 0.131 & 0.058 & 1.348 & \textbf{0.322} & \textbf{4.442} & \underline{0.608} \\
w/o PTS    & \underline{0.103} & \underline{0.040} & \underline{0.705} & 0.327 & 4.465 & 0.607 \\
C4D (Ours) & \textbf{0.103} & \textbf{0.040} & \textbf{0.705} & \underline{0.327} & \underline{4.459} & \textbf{0.609} \\
\bottomrule
\end{tabular}
\caption{\textbf{Ablation study} on the Sintel dataset.}
\label{tab:ablation}
% \vspace{-1em}
\end{table}

\begin{figure}[t]
\centering
\includegraphics[width=1.\linewidth]{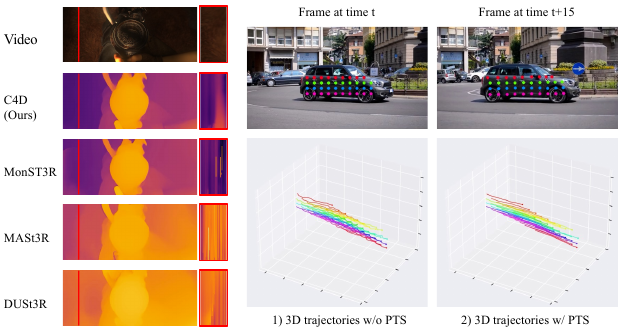}
\caption{\textbf{Ablation illustration of Point Trajectory Smoothness (PTS) objective.} The temporal depth and 3D trajectories become more smooth after applying PTS objective.}
\label{fig:ablation}
% \vspace{-3mm}
\end{figure}

\begin{figure}[t]
\centering
\includegraphics[width=1.\linewidth]{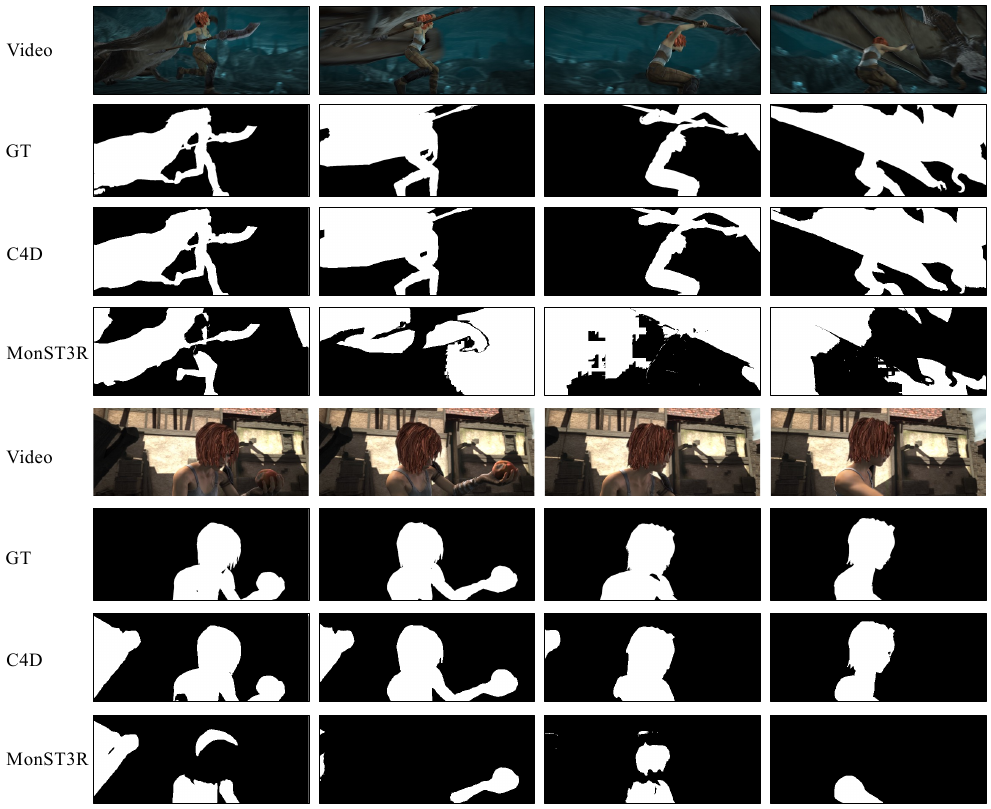}
\caption{\textbf{Qualitative comparison of motion mask on Sintel.} Our motion mask is more accurate than MonST3R's.}
\label{fig:mask}
% \vspace{-3mm}
\end{figure}

\subsection{Ablation Study}
\label{sec:ablation}
Ablation results in Table~\ref{tab:ablation} indicate that all loss functions are crucial. The proposed loss suite achieves the best pose estimation with minimal impact on video depth accuracy. Since the temporal smoothness of depth cannot be reflected by the quantitative metrics in Table~\ref{tab:ablation}, we show the temporal depth slice changes in Figure~\ref{fig:ablation}, following~\cite{yang2024depthv2, hu2024depthcrafter}, which demonstrates that our PTS objective is effective in producing more temporally smooth depth and 3D point trajectories. Note that while MonST3R also employs the CMA objective, the motion mask used in this objective is crucial, and our motion mask is more accurate than MonST3R's, as shown in Figure~\ref{fig:mask}. Due to page limitations, the ablation of DynPT is provided in the supplement.

\section{Conclusion}

In this paper, we introduce C4D, a framework for recovering 4D representations from monocular videos through joint prediction of dense pointmaps and temporal correspondences. 
Within this framework, a Dynamic-aware Point Tracker (DynPT), a correspondence-guided motion mask prediction, and correspondence-aided optimization are proposed to achieve accurate and smooth 4D reconstruction and camera pose estimation. Experiments demonstrate that C4D effectively reconstructs dynamic scenes, delivering competitive performance in depth estimation, camera pose estimation, and point tracking.

\section*{Acknowledgement}
This project is supported by the National Research Foundation, Singapore, under its Medium Sized Center for Advanced Robotics Technology Innovation.

{
    \small
    \bibliographystyle{ieeenat_fullname}
    \bibliography{main}

\begin{thebibliography}{74}
\providecommand{\natexlab}[1]{#1}
\providecommand{\url}[1]{\texttt{#1}}
\expandafter\ifx\csname urlstyle\endcsname\relax
  \providecommand{\doi}[1]{doi: #1}\else
  \providecommand{\doi}{doi: \begingroup \urlstyle{rm}\Url}\fi

\bibitem[Agarwal et~al.(2010)Agarwal, Snavely, Seitz, and Szeliski]{agarwal2010bundle}
Sameer Agarwal, Noah Snavely, Steven~M Seitz, and Richard Szeliski.
\newblock Bundle adjustment in the large.
\newblock In \emph{Computer Vision--ECCV 2010: 11th European Conference on Computer Vision, Heraklion, Crete, Greece, September 5-11, 2010, Proceedings, Part II 11}, pages 29--42. Springer, 2010.

\bibitem[Bay et~al.(2008)Bay, Ess, Tuytelaars, and Van~Gool]{bay2008speeded}
Herbert Bay, Andreas Ess, Tinne Tuytelaars, and Luc Van~Gool.
\newblock Speeded-up robust features (surf).
\newblock \emph{Computer vision and image understanding}, 110\penalty0 (3):\penalty0 346--359, 2008.

\bibitem[Butler et~al.(2012)Butler, Wulff, Stanley, and Black]{sintel}
Daniel~J. Butler, Jonas Wulff, Garrett~B. Stanley, and Michael~J. Black.
\newblock A naturalistic open source movie for optical flow evaluation.
\newblock In \emph{ECCV}, pages 611--625, 2012.

\bibitem[Chen et~al.(2024)Chen, Chen, Wang, and Pollefeys]{chen2024leap}
Weirong Chen, Le Chen, Rui Wang, and Marc Pollefeys.
\newblock {LEAP-VO}: Long-term effective any point tracking for visual odometry.
\newblock In \emph{CVPR}, pages 19844--19853, 2024.

\bibitem[Chen et~al.(2025)Chen, Chen, Xiu, Geiger, and Chen]{chen2025easi3r}
Xingyu Chen, Yue Chen, Yuliang Xiu, Andreas Geiger, and Anpei Chen.
\newblock Easi3r: Estimating disentangled motion from dust3r without training.
\newblock \emph{arXiv preprint arXiv:2503.24391}, 2025.

\bibitem[Cho et~al.(2025)Cho, Huang, Nam, An, Kim, and Lee]{locotrack}
Seokju Cho, Jiahui Huang, Jisu Nam, Honggyu An, Seungryong Kim, and Joon-Young Lee.
\newblock Local all-pair correspondence for point tracking.
\newblock In \emph{European Conference on Computer Vision}, pages 306--325. Springer, 2025.

\bibitem[Chu et~al.(2024)Chu, Ke, and Fragkiadaki]{chu2024dreamscene4d}
Wen-Hsuan Chu, Lei Ke, and Katerina Fragkiadaki.
\newblock Dreamscene4d: Dynamic multi-object scene generation from monocular videos.
\newblock \emph{arXiv preprint arXiv:2405.02280}, 2024.

\bibitem[Dai et~al.(2017)Dai, Chang, Savva, Halber, Funkhouser, and Nie{\ss}ner]{dai2017scannet}
Angela Dai, Angel~X. Chang, Manolis Savva, Maciej Halber, Thomas Funkhouser, and Matthias Nie{\ss}ner.
\newblock Scan{N}et: Richly-annotated 3{D} reconstructions of indoor scenes.
\newblock In \emph{CVPR}, pages 5828--5839, 2017.

\bibitem[Davison et~al.(2007)Davison, Reid, Molton, and Stasse]{davison2007monoslam}
Andrew~J Davison, Ian~D Reid, Nicholas~D Molton, and Olivier Stasse.
\newblock Monoslam: Real-time single camera slam.
\newblock \emph{IEEE transactions on pattern analysis and machine intelligence}, 29\penalty0 (6):\penalty0 1052--1067, 2007.

\bibitem[Doersch et~al.(2022)Doersch, Gupta, Markeeva, Recasens, Smaira, Aytar, Carreira, Zisserman, and Yang]{doersch2022tap}
Carl Doersch, Ankush Gupta, Larisa Markeeva, Adria Recasens, Lucas Smaira, Yusuf Aytar, Joao Carreira, Andrew Zisserman, and Yi Yang.
\newblock Tap-vid: A benchmark for tracking any point in a video.
\newblock \emph{Advances in Neural Information Processing Systems}, 35:\penalty0 13610--13626, 2022.

\bibitem[Doersch et~al.(2023)Doersch, Yang, Vecerik, Gokay, Gupta, Aytar, Carreira, and Zisserman]{doersch2023tapir}
Carl Doersch, Yi Yang, Mel Vecerik, Dilara Gokay, Ankush Gupta, Yusuf Aytar, Joao Carreira, and Andrew Zisserman.
\newblock Tapir: Tracking any point with per-frame initialization and temporal refinement.
\newblock In \emph{Proceedings of the IEEE/CVF International Conference on Computer Vision}, pages 10061--10072, 2023.

\bibitem[Dosovitskiy et~al.(2015)Dosovitskiy, Fischer, Ilg, Hausser, Hazirbas, Golkov, Van Der~Smagt, Cremers, and Brox]{dosovitskiy2015flownet}
Alexey Dosovitskiy, Philipp Fischer, Eddy Ilg, Philip Hausser, Caner Hazirbas, Vladimir Golkov, Patrick Van Der~Smagt, Daniel Cremers, and Thomas Brox.
\newblock Flownet: Learning optical flow with convolutional networks.
\newblock In \emph{Proceedings of the IEEE international conference on computer vision}, pages 2758--2766, 2015.

\bibitem[Duisterhof et~al.(2024)Duisterhof, Zust, Weinzaepfel, Leroy, Cabon, and Revaud]{duisterhof2024mast3r}
Bardienus Duisterhof, Lojze Zust, Philippe Weinzaepfel, Vincent Leroy, Yohann Cabon, and Jerome Revaud.
\newblock Mast3r-sfm: a fully-integrated solution for unconstrained structure-from-motion.
\newblock \emph{arXiv preprint arXiv:2409.19152}, 2024.

\bibitem[Geiger et~al.(2013)Geiger, Lenz, Stiller, and Urtasun]{geiger2013vision}
Andreas Geiger, Philip Lenz, Christoph Stiller, and Raquel Urtasun.
\newblock Vision meets robotics: The {KITTI} dataset.
\newblock 32\penalty0 (11):\penalty0 1231--1237, 2013.

\bibitem[Godard et~al.(2019)Godard, Mac~Aodha, Firman, and Brostow]{godard2019digging}
Cl{\'e}ment Godard, Oisin Mac~Aodha, Michael Firman, and Gabriel~J Brostow.
\newblock Digging into self-supervised monocular depth estimation.
\newblock In \emph{Proceedings of the IEEE/CVF international conference on computer vision}, pages 3828--3838, 2019.

\bibitem[Greff et~al.(2022)Greff, Belletti, Beyer, Doersch, Du, Duckworth, Fleet, Gnanapragasam, Golemo, Herrmann, et~al.]{greff2022kubric}
Klaus Greff, Francois Belletti, Lucas Beyer, Carl Doersch, Yilun Du, Daniel Duckworth, David~J Fleet, Dan Gnanapragasam, Florian Golemo, Charles Herrmann, et~al.
\newblock Kubric: A scalable dataset generator.
\newblock In \emph{Proceedings of the IEEE/CVF conference on computer vision and pattern recognition}, pages 3749--3761, 2022.

\bibitem[Han et~al.(2025)Han, An, Jung, Narihira, Seo, Fukuda, Kim, Hong, Mitsufuji, and Kim]{han2025d}
Jisang Han, Honggyu An, Jaewoo Jung, Takuya Narihira, Junyoung Seo, Kazumi Fukuda, Chaehyun Kim, Sunghwan Hong, Yuki Mitsufuji, and Seungryong Kim.
\newblock D\^{} 2ust3r: Enhancing 3d reconstruction with 4d pointmaps for dynamic scenes.
\newblock \emph{arXiv preprint arXiv:2504.06264}, 2025.

\bibitem[Harley et~al.(2022)Harley, Fang, and Fragkiadaki]{harley2022particle}
Adam~W Harley, Zhaoyuan Fang, and Katerina Fragkiadaki.
\newblock Particle video revisited: Tracking through occlusions using point trajectories.
\newblock In \emph{European Conference on Computer Vision}, pages 59--75. Springer, 2022.

\bibitem[Hartley and Zisserman(2003)]{hartley2003multiple}
Richard Hartley and Andrew Zisserman.
\newblock \emph{Multiple view geometry in computer vision}.
\newblock Cambridge university press, 2003.

\bibitem[He et~al.(2016)He, Zhang, Ren, and Sun]{resnet}
Kaiming He, Xiangyu Zhang, Shaoqing Ren, and Jian Sun.
\newblock Deep residual learning for image recognition.
\newblock In \emph{Proceedings of the IEEE conference on computer vision and pattern recognition}, pages 770--778, 2016.

\bibitem[Hu et~al.(2024)Hu, Gao, Li, Zhao, Cun, Zhang, Quan, and Shan]{hu2024depthcrafter}
Wenbo Hu, Xiangjun Gao, Xiaoyu Li, Sijie Zhao, Xiaodong Cun, Yong Zhang, Long Quan, and Ying Shan.
\newblock Depth{C}rafter: Generating consistent long depth sequences for open-world videos.
\newblock \emph{arXiv preprint arXiv:2409.02095}, 2024.

\bibitem[Kappel et~al.(2024)Kappel, Hahlbohm, Scholz, Castillo, Theobalt, Eisemann, Golyanik, and Magnor]{kappel2024d}
Moritz Kappel, Florian Hahlbohm, Timon Scholz, Susana Castillo, Christian Theobalt, Martin Eisemann, Vladislav Golyanik, and Marcus Magnor.
\newblock D-npc: Dynamic neural point clouds for non-rigid view synthesis from monocular video.
\newblock \emph{arXiv preprint arXiv:2406.10078}, 2024.

\bibitem[Karaev et~al.(2023)Karaev, Rocco, Graham, Neverova, Vedaldi, and Rupprecht]{karaev2023cotracker}
Nikita Karaev, Ignacio Rocco, Benjamin Graham, Natalia Neverova, Andrea Vedaldi, and Christian Rupprecht.
\newblock Cotracker: It is better to track together.
\newblock \emph{arXiv preprint arXiv:2307.07635}, 2023.

\bibitem[Karaev et~al.(2024)Karaev, Makarov, Wang, Neverova, Vedaldi, and Rupprecht]{karaev2024cotracker3}
Nikita Karaev, Iurii Makarov, Jianyuan Wang, Natalia Neverova, Andrea Vedaldi, and Christian Rupprecht.
\newblock Cotracker3: Simpler and better point tracking by pseudo-labelling real videos.
\newblock \emph{arXiv preprint arXiv:2410.11831}, 2024.

\bibitem[Ke et~al.(2024)Ke, Obukhov, Huang, Metzger, Daudt, and Schindler]{ke2024repurposing}
Bingxin Ke, Anton Obukhov, Shengyu Huang, Nando Metzger, Rodrigo~Caye Daudt, and Konrad Schindler.
\newblock Repurposing diffusion-based image generators for monocular depth estimation.
\newblock In \emph{Proceedings of the IEEE/CVF Conference on Computer Vision and Pattern Recognition}, pages 9492--9502, 2024.

\bibitem[Kerbl et~al.(2023)Kerbl, Kopanas, Leimk{\"u}hler, and Drettakis]{kerbl20233d}
Bernhard Kerbl, Georgios Kopanas, Thomas Leimk{\"u}hler, and George Drettakis.
\newblock 3d gaussian splatting for real-time radiance field rendering.
\newblock \emph{ACM Trans. Graph.}, 42\penalty0 (4):\penalty0 139--1, 2023.

\bibitem[Kong et~al.(2025{\natexlab{a}})Kong, Yang, and Wang]{kong2025efficient}
Hanyang Kong, Xingyi Yang, and Xinchao Wang.
\newblock Efficient gaussian splatting for monocular dynamic scene rendering via sparse time-variant attribute modeling.
\newblock In \emph{Proceedings of the AAAI Conference on Artificial Intelligence}, pages 4374--4382, 2025{\natexlab{a}}.

\bibitem[Kong et~al.(2025{\natexlab{b}})Kong, Yang, and Wang]{kong2025generative}
Hanyang Kong, Xingyi Yang, and Xinchao Wang.
\newblock Generative sparse-view gaussian splatting.
\newblock In \emph{Proceedings of the Computer Vision and Pattern Recognition Conference}, pages 26745--26755, 2025{\natexlab{b}}.

\bibitem[Kong et~al.(2025{\natexlab{c}})Kong, Yang, and Wang]{rogsplat}
Hanyang Kong, Xingyi Yang, and Xinchao Wang.
\newblock Rogsplat: Robust gaussian splatting via generative priors.
\newblock In \emph{Proceedings of the IEEE International Conference on Computer Vision}, 2025{\natexlab{c}}.

\bibitem[Kopf et~al.(2021)Kopf, Rong, and Huang]{kopf2021robust}
Johannes Kopf, Xuejian Rong, and Jia-Bin Huang.
\newblock Robust consistent video depth estimation.
\newblock In \emph{Proceedings of the IEEE/CVF Conference on Computer Vision and Pattern Recognition}, pages 1611--1621, 2021.

\bibitem[Lei et~al.(2024)Lei, Weng, Harley, Guibas, and Daniilidis]{lei2024mosca}
Jiahui Lei, Yijia Weng, Adam Harley, Leonidas Guibas, and Kostas Daniilidis.
\newblock Mosca: Dynamic gaussian fusion from casual videos via 4d motion scaffolds.
\newblock \emph{arXiv preprint arXiv:2405.17421}, 2024.

\bibitem[Leroy et~al.(2024)Leroy, Cabon, and Revaud]{leroy2024mast3r}
Vincent Leroy, Yohann Cabon, and J{\'e}r{\^o}me Revaud.
\newblock Grounding image matching in 3d with mast3r.
\newblock \emph{arXiv preprint arXiv:2406.09756}, 2024.

\bibitem[Liu et~al.(2024)Liu, Liu, Wang, Lv, Wang, Wang, and Hou]{liu2024modgs}
Qingming Liu, Yuan Liu, Jiepeng Wang, Xianqiang Lv, Peng Wang, Wenping Wang, and Junhui Hou.
\newblock Modgs: Dynamic gaussian splatting from causually-captured monocular videos.
\newblock \emph{arXiv preprint arXiv:2406.00434}, 2024.

\bibitem[Lowe(1999)]{lowe1999object}
David~G Lowe.
\newblock Object recognition from local scale-invariant features.
\newblock In \emph{Proceedings of the seventh IEEE international conference on computer vision}, pages 1150--1157. Ieee, 1999.

\bibitem[Lowe(2004)]{lowe2004distinctive}
David~G Lowe.
\newblock Distinctive image features from scale-invariant keypoints.
\newblock \emph{International journal of computer vision}, 60:\penalty0 91--110, 2004.

\bibitem[Mayer et~al.(2016)Mayer, Ilg, Hausser, Fischer, Cremers, Dosovitskiy, and Brox]{mayer2016large}
Nikolaus Mayer, Eddy Ilg, Philip Hausser, Philipp Fischer, Daniel Cremers, Alexey Dosovitskiy, and Thomas Brox.
\newblock A large dataset to train convolutional networks for disparity, optical flow, and scene flow estimation.
\newblock In \emph{Proceedings of the IEEE conference on computer vision and pattern recognition}, pages 4040--4048, 2016.

\bibitem[Mur-Artal et~al.(2015)Mur-Artal, Montiel, and Tardos]{mur2015orb}
Raul Mur-Artal, Jose Maria~Martinez Montiel, and Juan~D Tardos.
\newblock Orb-slam: A versatile and accurate monocular slam system.
\newblock \emph{IEEE transactions on robotics}, 31\penalty0 (5):\penalty0 1147--1163, 2015.

\bibitem[Neoral et~al.(2024)Neoral, {\v{S}}er{\`y}ch, and Matas]{neoral2024mft}
Michal Neoral, Jon{\'a}{\v{s}} {\v{S}}er{\`y}ch, and Ji{\v{r}}{\'\i} Matas.
\newblock Mft: Long-term tracking of every pixel.
\newblock In \emph{Proceedings of the IEEE/CVF Winter Conference on Applications of Computer Vision}, pages 6837--6847, 2024.

\bibitem[Newcombe et~al.(2011)Newcombe, Lovegrove, and Davison]{newcombe2011dtam}
Richard~A Newcombe, Steven~J Lovegrove, and Andrew~J Davison.
\newblock Dtam: Dense tracking and mapping in real-time.
\newblock In \emph{2011 international conference on computer vision}, pages 2320--2327. IEEE, 2011.

\bibitem[Palazzolo et~al.(2019)Palazzolo, Behley, Lottes, Giguere, and Stachniss]{palazzolo2019refusion}
Emanuele Palazzolo, Jens Behley, Philipp Lottes, Philippe Giguere, and Cyrill Stachniss.
\newblock Refusion: 3d reconstruction in dynamic environments for {RGB-D} cameras exploiting residuals.
\newblock pages 7855--7862, 2019.

\bibitem[Perazzi et~al.(2016)Perazzi, Pont-Tuset, McWilliams, Van~Gool, Gross, and Sorkine-Hornung]{perazzi2016benchmark}
Federico Perazzi, Jordi Pont-Tuset, Brian McWilliams, Luc Van~Gool, Markus Gross, and Alexander Sorkine-Hornung.
\newblock A benchmark dataset and evaluation methodology for video object segmentation.
\newblock In \emph{Proceedings of the IEEE conference on computer vision and pattern recognition}, pages 724--732, 2016.

\bibitem[Ravi et~al.(2024)Ravi, Gabeur, Hu, Hu, Ryali, Ma, Khedr, R{\"a}dle, Rolland, Gustafson, et~al.]{ravi2024sam}
Nikhila Ravi, Valentin Gabeur, Yuan-Ting Hu, Ronghang Hu, Chaitanya Ryali, Tengyu Ma, Haitham Khedr, Roman R{\"a}dle, Chloe Rolland, Laura Gustafson, et~al.
\newblock Sam 2: Segment anything in images and videos.
\newblock \emph{arXiv preprint arXiv:2408.00714}, 2024.

\bibitem[Rousseeuw(1984)]{rousseeuw1984least}
Peter~J Rousseeuw.
\newblock Least median of squares regression.
\newblock \emph{Journal of the American statistical association}, 79\penalty0 (388):\penalty0 871--880, 1984.

\bibitem[Rublee et~al.(2011)Rublee, Rabaud, Konolige, and Bradski]{rublee2011orb}
Ethan Rublee, Vincent Rabaud, Kurt Konolige, and Gary Bradski.
\newblock Orb: An efficient alternative to sift or surf.
\newblock In \emph{2011 International conference on computer vision}, pages 2564--2571. Ieee, 2011.

\bibitem[Sameer(2009)]{sameer2009building}
Agarwal Sameer.
\newblock Building rome in a day.
\newblock \emph{Proc. ICCV, 2009}, 2009.

\bibitem[Schonberger and Frahm(2016)]{schonberger2016structure}
Johannes~L Schonberger and Jan-Michael Frahm.
\newblock Structure-from-motion revisited.
\newblock In \emph{Proceedings of the IEEE conference on computer vision and pattern recognition}, pages 4104--4113, 2016.

\bibitem[Shao et~al.(2024)Shao, Yang, Zhou, Zhang, Shen, Poggi, and Liao]{shao2024learning}
Jiahao Shao, Yuanbo Yang, Hongyu Zhou, Youmin Zhang, Yujun Shen, Matteo Poggi, and Yiyi Liao.
\newblock Learning temporally consistent video depth from video diffusion priors.
\newblock \emph{arXiv preprint arXiv:2406.01493}, 2024.

\bibitem[Smith and Topin(2019)]{smith2019super}
Leslie~N Smith and Nicholay Topin.
\newblock Super-convergence: Very fast training of neural networks using large learning rates.
\newblock In \emph{Artificial intelligence and machine learning for multi-domain operations applications}, pages 369--386. SPIE, 2019.

\bibitem[Stearns et~al.(2024)Stearns, Harley, Uy, Dubost, Tombari, Wetzstein, and Guibas]{stearns2024dynamic}
Colton Stearns, Adam Harley, Mikaela Uy, Florian Dubost, Federico Tombari, Gordon Wetzstein, and Leonidas Guibas.
\newblock Dynamic gaussian marbles for novel view synthesis of casual monocular videos.
\newblock \emph{arXiv preprint arXiv:2406.18717}, 2024.

\bibitem[Steinbr{\"u}cker et~al.(2011)Steinbr{\"u}cker, Sturm, and Cremers]{steinbrucker2011real}
Frank Steinbr{\"u}cker, J{\"u}rgen Sturm, and Daniel Cremers.
\newblock Real-time visual odometry from dense rgb-d images.
\newblock In \emph{2011 IEEE international conference on computer vision workshops (ICCV Workshops)}, pages 719--722. IEEE, 2011.

\bibitem[Sturm et~al.(2012)Sturm, Engelhard, Endres, Burgard, and Cremers]{tum}
J{\"u}rgen Sturm, Nikolas Engelhard, Felix Endres, Wolfram Burgard, and Daniel Cremers.
\newblock A benchmark for the evaluation of {RGB-D SLAM} systems.
\newblock pages 573--580, 2012.

\bibitem[Sucar et~al.(2025)Sucar, Lai, Insafutdinov, and Vedaldi]{sucar2025dynamic}
Edgar Sucar, Zihang Lai, Eldar Insafutdinov, and Andrea Vedaldi.
\newblock Dynamic point maps: A versatile representation for dynamic 3d reconstruction.
\newblock \emph{arXiv preprint arXiv:2503.16318}, 2025.

\bibitem[Sun et~al.(2018)Sun, Yang, Liu, and Kautz]{sun2018pwc}
Deqing Sun, Xiaodong Yang, Ming-Yu Liu, and Jan Kautz.
\newblock Pwc-net: Cnns for optical flow using pyramid, warping, and cost volume.
\newblock In \emph{Proceedings of the IEEE conference on computer vision and pattern recognition}, pages 8934--8943, 2018.

\bibitem[Sun et~al.(2024)Sun, Huang, Ma, Lyu, Cao, and Qi]{sun2024splatter}
Yang-Tian Sun, Yihua Huang, Lin Ma, Xiaoyang Lyu, Yan-Pei Cao, and Xiaojuan Qi.
\newblock Splatter a video: Video gaussian representation for versatile processing.
\newblock \emph{Advances in Neural Information Processing Systems}, 37:\penalty0 50401--50425, 2024.

\bibitem[Teed and Deng(2020)]{teed2020raft}
Zachary Teed and Jia Deng.
\newblock Raft: Recurrent all-pairs field transforms for optical flow.
\newblock In \emph{Computer Vision--ECCV 2020: 16th European Conference, Glasgow, UK, August 23--28, 2020, Proceedings, Part II 16}, pages 402--419. Springer, 2020.

\bibitem[Teed and Deng(2021)]{teed2021droid}
Zachary Teed and Jia Deng.
\newblock Droid-slam: Deep visual slam for monocular, stereo, and rgb-d cameras.
\newblock \emph{Advances in neural information processing systems}, 34:\penalty0 16558--16569, 2021.

\bibitem[Teed et~al.(2024)Teed, Lipson, and Deng]{teed2024deep}
Zachary Teed, Lahav Lipson, and Jia Deng.
\newblock Deep patch visual odometry.
\newblock \emph{Advances in Neural Information Processing Systems}, 36, 2024.

\bibitem[Triggs et~al.(2000)Triggs, McLauchlan, Hartley, and Fitzgibbon]{triggs2000bundle}
Bill Triggs, Philip~F McLauchlan, Richard~I Hartley, and Andrew~W Fitzgibbon.
\newblock Bundle adjustment—a modern synthesis.
\newblock In \emph{Vision Algorithms: Theory and Practice: International Workshop on Vision Algorithms Corfu, Greece, September 21--22, 1999 Proceedings}, pages 298--372. Springer, 2000.

\bibitem[Wang et~al.(2023{\natexlab{a}})Wang, Chang, Cai, Li, Hariharan, Holynski, and Snavely]{wang2023tracking}
Qianqian Wang, Yen-Yu Chang, Ruojin Cai, Zhengqi Li, Bharath Hariharan, Aleksander Holynski, and Noah Snavely.
\newblock Tracking everything everywhere all at once.
\newblock In \emph{Proceedings of the IEEE/CVF International Conference on Computer Vision}, pages 19795--19806, 2023{\natexlab{a}}.

\bibitem[Wang et~al.(2024{\natexlab{a}})Wang, Ye, Gao, Austin, Li, and Kanazawa]{wang2024shape}
Qianqian Wang, Vickie Ye, Hang Gao, Jake Austin, Zhengqi Li, and Angjoo Kanazawa.
\newblock Shape of motion: 4d reconstruction from a single video.
\newblock \emph{arXiv preprint arXiv:2407.13764}, 2024{\natexlab{a}}.

\bibitem[Wang et~al.(2024{\natexlab{b}})Wang, Leroy, Cabon, Chidlovskii, and Revaud]{wang2024dust3r}
Shuzhe Wang, Vincent Leroy, Yohann Cabon, Boris Chidlovskii, and Jerome Revaud.
\newblock Dust3r: Geometric 3d vision made easy.
\newblock In \emph{Proceedings of the IEEE/CVF Conference on Computer Vision and Pattern Recognition}, pages 20697--20709, 2024{\natexlab{b}}.

\bibitem[Wang et~al.(2024{\natexlab{c}})Wang, Yang, Shen, Jiang, and Wang]{wang2024gflow}
Shizun Wang, Xingyi Yang, Qiuhong Shen, Zhenxiang Jiang, and Xinchao Wang.
\newblock Gflow: Recovering 4d world from monocular video.
\newblock \emph{arXiv preprint arXiv:2405.18426}, 2024{\natexlab{c}}.

\bibitem[Wang et~al.(2023{\natexlab{b}})Wang, Shi, Li, Huang, Cao, Zhang, Xian, and Lin]{wang2023neural}
Yiran Wang, Min Shi, Jiaqi Li, Zihao Huang, Zhiguo Cao, Jianming Zhang, Ke Xian, and Guosheng Lin.
\newblock Neural video depth stabilizer.
\newblock In \emph{Proceedings of the IEEE/CVF International Conference on Computer Vision}, pages 9466--9476, 2023{\natexlab{b}}.

\bibitem[Wang et~al.(2025)Wang, Lipson, and Deng]{wang2025searaft}
Yihan Wang, Lahav Lipson, and Jia Deng.
\newblock Sea-raft: Simple, efficient, accurate raft for optical flow.
\newblock In \emph{European Conference on Computer Vision}, pages 36--54. Springer, 2025.

\bibitem[Weinzaepfel et~al.(2023)Weinzaepfel, Lucas, Leroy, Cabon, Arora, Br{\'e}gier, Csurka, Antsfeld, Chidlovskii, and Revaud]{weinzaepfel2023croco}
Philippe Weinzaepfel, Thomas Lucas, Vincent Leroy, Yohann Cabon, Vaibhav Arora, Romain Br{\'e}gier, Gabriela Csurka, Leonid Antsfeld, Boris Chidlovskii, and J{\'e}r{\^o}me Revaud.
\newblock Croco v2: Improved cross-view completion pre-training for stereo matching and optical flow.
\newblock In \emph{Proceedings of the IEEE/CVF International Conference on Computer Vision}, pages 17969--17980, 2023.

\bibitem[Wu(2013)]{wu2013towards}
Changchang Wu.
\newblock Towards linear-time incremental structure from motion.
\newblock In \emph{2013 International Conference on 3D Vision-3DV 2013}, pages 127--134. IEEE, 2013.

\bibitem[Xie et~al.(2024)Xie, Yang, Xie, and Zisserman]{xie2024moving}
Junyu Xie, Charig Yang, Weidi Xie, and Andrew Zisserman.
\newblock Moving object segmentation: All you need is sam (and flow).
\newblock In \emph{Proceedings of the Asian Conference on Computer Vision}, pages 162--178, 2024.

\bibitem[Xu et~al.(2022)Xu, Zhang, Cai, Rezatofighi, and Tao]{xu2022gmflow}
Haofei Xu, Jing Zhang, Jianfei Cai, Hamid Rezatofighi, and Dacheng Tao.
\newblock Gmflow: Learning optical flow via global matching.
\newblock In \emph{Proceedings of the IEEE/CVF conference on computer vision and pattern recognition}, pages 8121--8130, 2022.

\bibitem[Yang et~al.(2024)Yang, Kang, Huang, Zhao, Xu, Feng, and Zhao]{yang2024depthv2}
Lihe Yang, Bingyi Kang, Zilong Huang, Zhen Zhao, Xiaogang Xu, Jiashi Feng, and Hengshuang Zhao.
\newblock Depth anything {V}2.
\newblock \emph{arXiv preprint arXiv:2406.09414}, 2024.

\bibitem[Yin et~al.(2021)Yin, Zhang, Wang, Niklaus, Mai, Chen, and Shen]{yin2021learning}
Wei Yin, Jianming Zhang, Oliver Wang, Simon Niklaus, Long Mai, Simon Chen, and Chunhua Shen.
\newblock Learning to recover 3d scene shape from a single image.
\newblock In \emph{Proceedings of the IEEE/CVF Conference on Computer Vision and Pattern Recognition}, pages 204--213, 2021.

\bibitem[Zhang et~al.(2024)Zhang, Herrmann, Hur, Jampani, Darrell, Cole, Sun, and Yang]{zhang2024monst3r}
Junyi Zhang, Charles Herrmann, Junhwa Hur, Varun Jampani, Trevor Darrell, Forrester Cole, Deqing Sun, and Ming-Hsuan Yang.
\newblock Monst3r: A simple approach for estimating geometry in the presence of motion.
\newblock \emph{arXiv preprint arXiv:2410.03825}, 2024.

\bibitem[Zhang et~al.(2021)Zhang, Cole, Tucker, Freeman, and Dekel]{zhang2021consistent}
Zhoutong Zhang, Forrester Cole, Richard Tucker, William~T Freeman, and Tali Dekel.
\newblock Consistent depth of moving objects in video.
\newblock \emph{ACM Transactions on Graphics (ToG)}, 40\penalty0 (4):\penalty0 1--12, 2021.

\bibitem[Zhang et~al.(2022)Zhang, Cole, Li, Rubinstein, Snavely, and Freeman]{zhang2022structure}
Zhoutong Zhang, Forrester Cole, Zhengqi Li, Michael Rubinstein, Noah Snavely, and William~T Freeman.
\newblock Structure and motion from casual videos.
\newblock In \emph{European Conference on Computer Vision}, pages 20--37. Springer, 2022.

\bibitem[Zhao et~al.(2022)Zhao, Liu, Guo, Wang, and Liu]{zhao2022particlesfm}
Wang Zhao, Shaohui Liu, Hengkai Guo, Wenping Wang, and Yong-Jin Liu.
\newblock Particlesfm: Exploiting dense point trajectories for localizing moving cameras in the wild.
\newblock In \emph{European Conference on Computer Vision}, pages 523--542. Springer, 2022.

\end{thebibliography}
}

\clearpage
\setcounter{page}{1}
\maketitlesupplementary

\section{More Visual Results}
\subsection{4D Reconstruction}
\begin{figure*}[t]
\includegraphics[width=\linewidth]{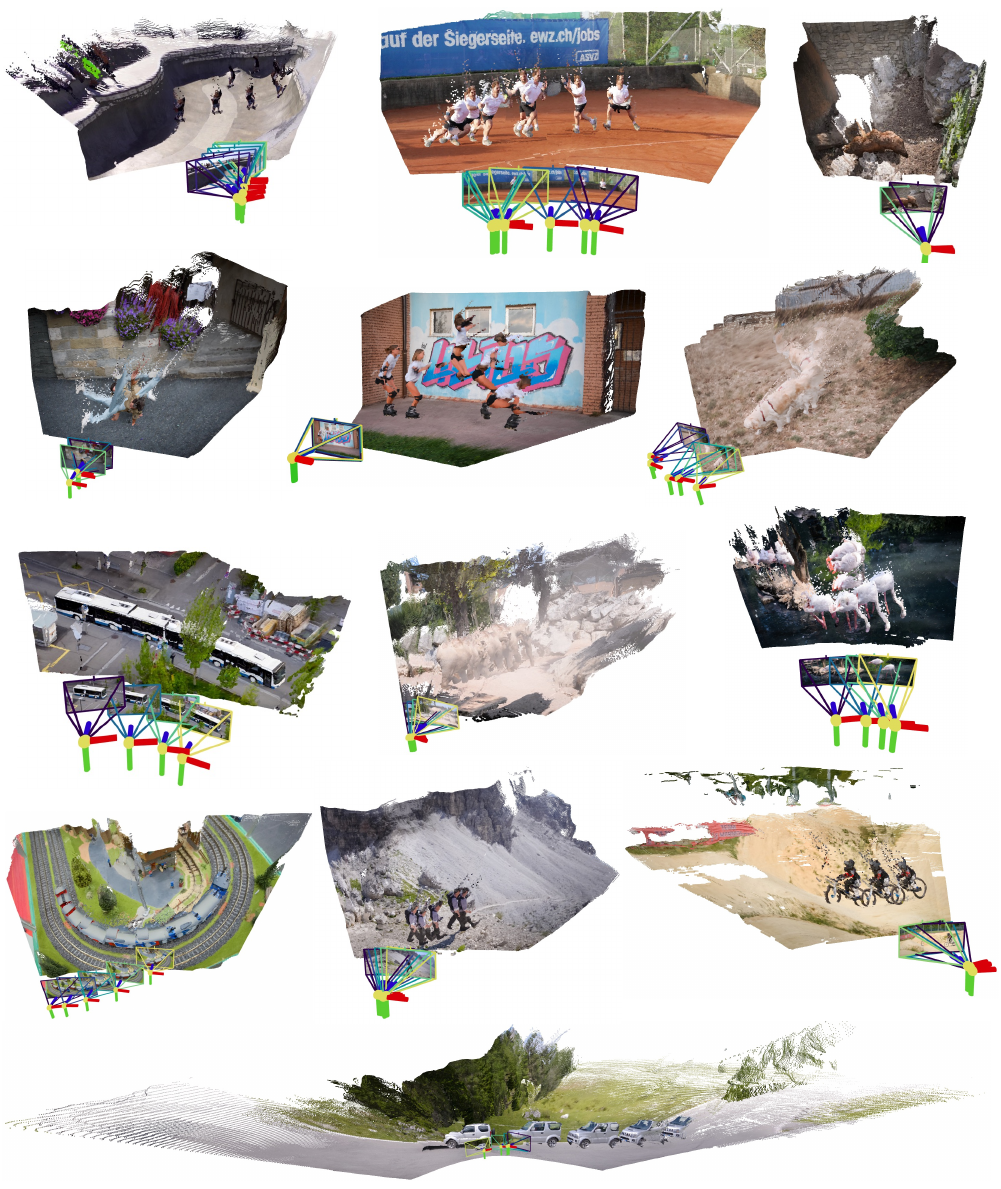}
\caption{\textbf{Visual results of 4D reconstruction on DAVIS dataset~\cite{perazzi2016benchmark}.} C4D can reconstruct the dynamic scene and recover camera parameters from monocular video input.}
\label{fig:4d}
\newpage
\end{figure*}

Given only a monocular video as input, C4D can reconstruct dynamic scenes along with camera parameters. Visual results are shown in Figure~\ref{fig:4d}. To provide a comprehensive view of the 4D reconstruction, the static regions across all frames are retained, while the dynamic regions from uniformly sampled frames are also displayed.

\begin{figure*}[t]
\includegraphics[width=\linewidth]{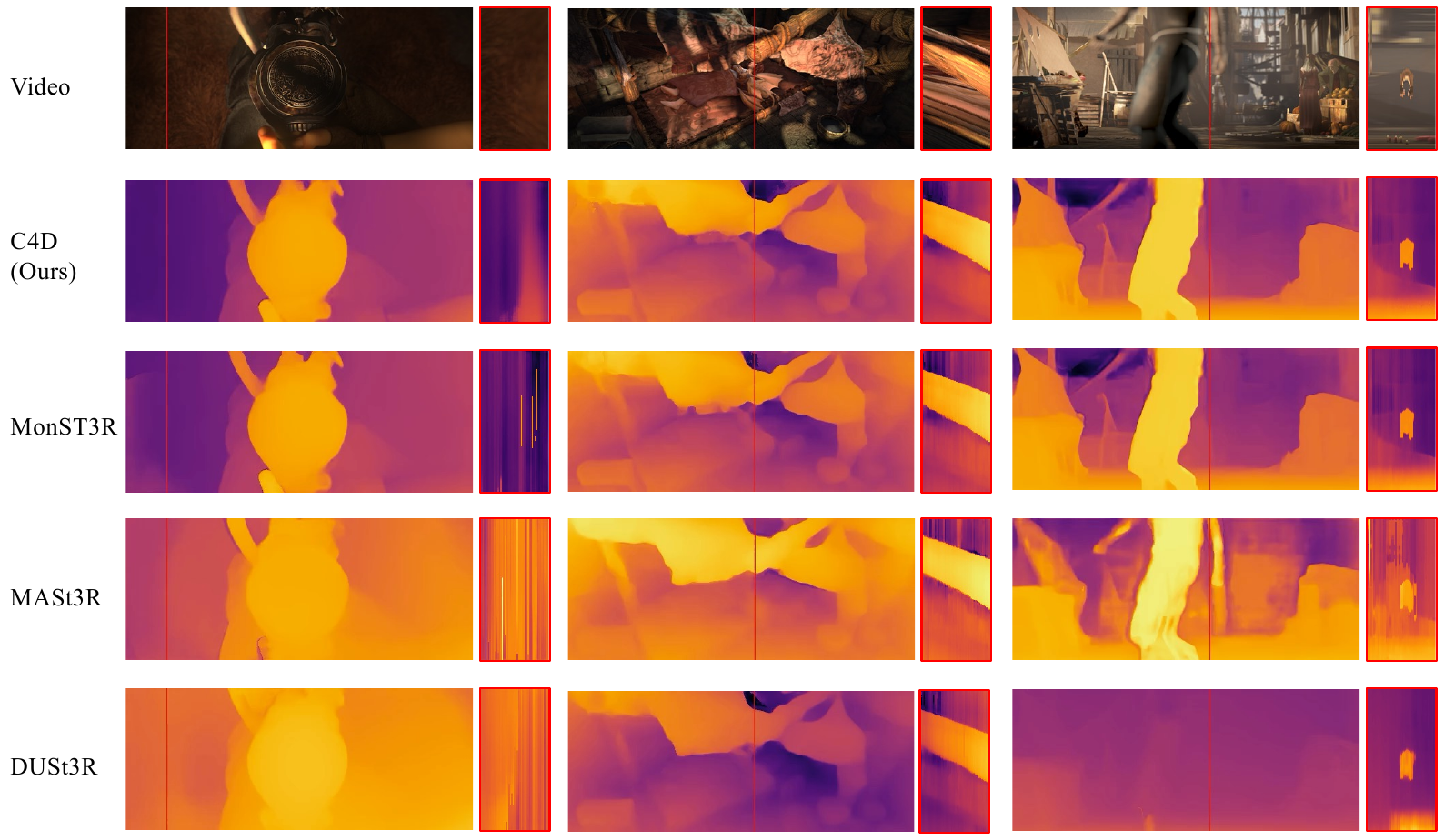}
\caption{\textbf{Qualitative comparison of video depth on Sintel~\cite{sintel}.} We compare C4D with MonST3R, MASt3R, and DUSt3R. To better visualize the temporal depth quality, we highlight $y$-$t$ depth slices along the vertical red line within red boxes. For optimal viewing, please zoom in.}
\label{fig:depth}
\end{figure*}

\begin{figure*}[t]
\includegraphics[width=\linewidth]{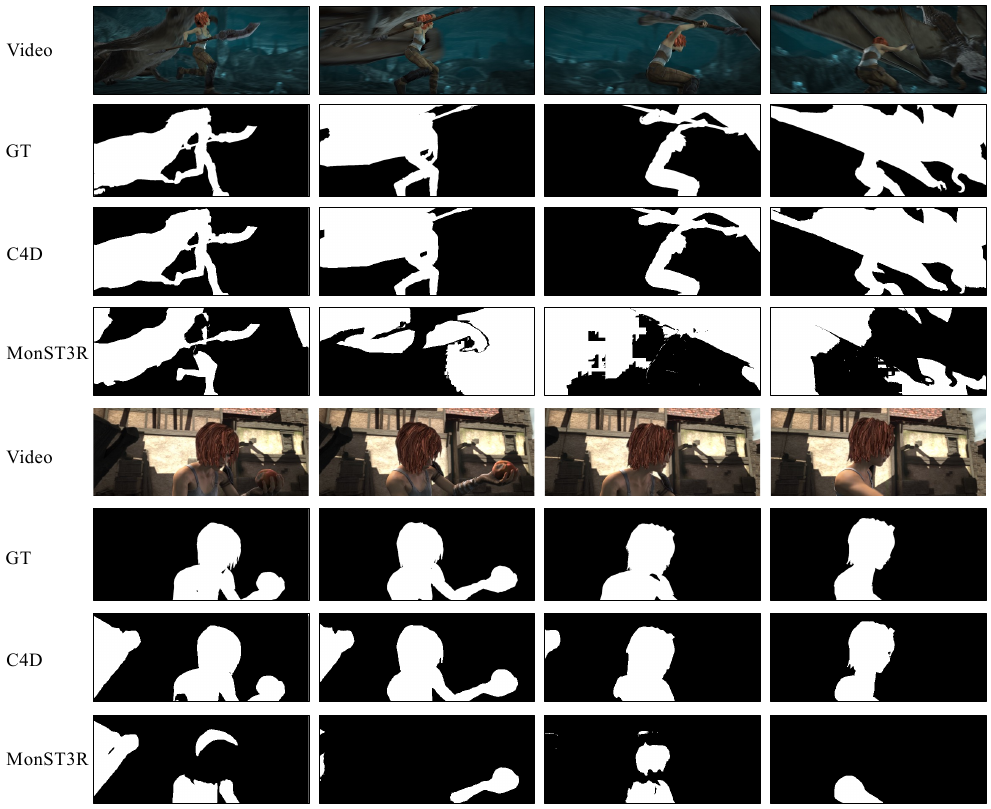}
\caption{\textbf{Qualitative comparison of motion mask on Sintel dataset~\cite{sintel}.} We present the motion masks generated by C4D and MonST3R. Video frames and ground-truth motion masks are also included for reference. The white regions indicate dynamic areas.}
\label{fig:motion}
\end{figure*}

\subsection{Temporally Smooth Video Depth}
In Figure~\ref{fig:depth}, we compare the video depth estimation results of C4D with other DUSt3R-based methods, including DUSt3R~\cite{wang2024dust3r}, MASt3R~\cite{leroy2024mast3r}, and MonST3R~\cite{zhang2024monst3r}. In addition to producing more accurate depth, C4D demonstrates superior temporal smoothness compared to other methods. To illustrate this, we highlight the $y$-$t$ depth slices along the vertical red line in the red boxes, showing the depth variation over time. As observed, C4D achieves temporally smoother depth results, thanks to the \textit{Point Trajectory Smoothness (PTS)} objective. In contrast, other methods exhibit zigzag artifacts in the $y$-$t$ depth slices, indicating flickering artifacts in the video depth.

\subsection{Motion Mask}
One of the most critical aspects of reconstructing dynamic scenes is identifying dynamic regions, that is, predicting motion masks. In Figure~\ref{fig:motion}, we provide a qualitative comparison of motion masks generated by our C4D and the concurrent work MonST3R on the Sintel dataset~\cite{sintel}. This dataset poses significant challenges due to its fast camera motion, large object motion, and motion blur.

In the first video, C4D demonstrates its ability to generate reliable motion masks on the Sintel dataset, even when a large portion of the frame content is dynamic. This success is attributed to our proposed correspondence-guided motion mask prediction strategy. In contrast, MonST3R struggles to recognize such dynamic regions under these challenging conditions.
In the second video, C4D predicts more complete motion masks, whereas MonST3R only generates partial results. This improvement is due to C4D's consideration of multi-frame motion cues in our motion mask prediction strategy, which is crucial for practical scenarios.

\section{More Experimental Results}
\begin{table}[th]
    % \vspace{-1em}
    \centering
    % \footnotesize
    \renewcommand{\arraystretch}{1.2}
    \begin{tabular}{c|ccc}
        Method & Ours & MonST3R~\cite{zhang2024monst3r} & FlowP-SAM~\cite{xie2024moving} \\
        \cline{1-4}
        % \cmidrule
        $IoU \uparrow$ & 47.89 & 31.57 & 42.23 \\
    \end{tabular}
    % \vspace{-1.2em}
    \caption{Motion segmentation results on DAVIS2016. Note that the evaluation is conducted without Hungarian matching between predicted and ground-truth motion masks.}
    \label{tab:motion-seg}
    % \vspace{-2.5em}
\end{table}

\subsection{Motion Segmentation Results}
Unlike prompt-based video segmentation like SAM2~\cite{ravi2024sam}, motion segmentation aims to automatically segment the moving regions in the video. We evaluate our method on DAVIS 2016~\cite{perazzi2016benchmark} and compare it with some automatic motion segmentation methods in Tab.~\ref{tab:motion-seg}, where our approach outperforms both  MonST3R~\cite{zhang2024monst3r} and the state-of-the-art supervised method, FlowP-SAM~\cite{xie2024moving}.
Note that the evaluation is conducted without Hungarian matching between predicted and ground-truth motion masks.

\subsection{Ablation on Different Tracker Variants}
The tracker needs to predict additional mobility to infer the dynamic mask, which is a more difficult learning problem as it requires understanding spatial relationships. Table.~\ref{tab:tracking_rebuttal} shows that a sole CNN or 3D-aware encoder struggles with multi-tasking, whereas combining both improves performance.

\begin{table*}
\centering

\begin{tabular}{lccccccccc}
\toprule
\multirow{2}{*}{Method} & 
MOVi-E & Pan. MOVi-E & MOVi-F & \multicolumn{3}{c}{TAP-Vid DAVIS} & \multicolumn{3}{c}{TAP-Vid Kinetics} \\ 
\cmidrule(lr){2-4} \cmidrule(lr){5-7} \cmidrule(lr){8-10} 
    & D-ACC~$\uparrow$  & D-ACC~$\uparrow$ &  D-ACC~$\uparrow$  & AJ~$\uparrow$ & $<\delta^x_\textrm{avg}$~$\uparrow$ & OA~$\uparrow$ &  AJ~$\uparrow$ & $<\delta^x_\textrm{avg}$~$\uparrow$ & OA~$\uparrow$  \\ 
\midrule
DynPT & \textbf{87.9} & \textbf{94.1} & \textbf{91.5}  & \textbf{61.6}  & \textbf{75.4}  & \textbf{87.4}  & \textbf{47.8}  & \textbf{62.6}  &  \textbf{82.3} \\
CE only & 82.6 &  90.4  &  86.8  & 60.6  &  74.3  &  86.8  & 46.2  &  62.1  &  81.7 \\
3E only & 85.4  & 92.2  &  90.4 &  42.4  &  56.6  &  73.4  & 38.9  &  54.6  &  70.4 \\
\bottomrule
\end{tabular}
% \vspace{-1em}
\caption{
Ablation study of different design choices for DynPT. ``CE" denotes the use of a CNN encoder, while ``3E" refers to the 3D-aware encoder.
}
\label{tab:tracking_rebuttal}
% \vspace{-2em}
% \vspace{-1.5em}
\end{table*}

\section{More Technical Details}

\subsection{Dynamic-aware Point Tracker (DynPT)}
The ground truth used to supervise confidence is defined by an indicator that determines whether the predicted position lies within 12 pixels of the ground truth position.
And since there are no currently available labels for mobility, we use the rich annotations provided by the Kubric~\cite{greff2022kubric} generator to generate ground-truth mobility labels. Specifically, the ``positions" label, which describes the position of an object for each frame in world coordinates, is utilized. 

As the movements of objects in Kubric are entirely rigid, we determine an object's mobility as follows: if the temporal difference in the "position" of an object exceeds a predefined threshold (e.g., 0.01), all the tracking points associated with that object are labeled as dynamic (i.e., mobility is labeled as True). 

It is important to note that although an ``is\_dynamic" label is provided in Kubric, it only indicates whether the object is stationary on the floor (False) or being tossed (True) at the initial frame. However, some objects may collide and move in subsequent frames. In such cases, the ``is\_dynamic" label does not accurately represent the object's mobility, necessitating the use of our threshold-based approach.

We train DynPT on the training sets of the panning MOVi-E and MOVi-F datasets. These datasets are chosen for their non-trivial camera movements and motion blur, which closely resemble real-world scenarios. For evaluation, in addition to the the panning MOVi-E and MOVi-F datasets, we also evaluate on the MOVi-E dataset to assess the generalization ability of DynPT. 

During inference, DynPT processes videos by querying a sparse set of points in a sliding window manner to maintain computational efficiency as in~\cite{karaev2023cotracker, karaev2024cotracker3}. The query points are sampled based on grids: the image is divided into grids of $20\times20$ pixels, and one point with the maximum image gradient is sampled from each grid to capture the most distinguishable descriptor. Additionally, one random point is sampled from each grid to ensure diversity and prevent bias towards only high-gradient areas. This combination of gradient-based sampling and random sampling ensures a balanced selection of points, enabling robust and diverse feature extraction across the image.

\subsection{Point Trajectory Smoothness (PTS) Objective}
The primary goal of this objective is to ensure temporal smoothness in the per-frame pointmaps. Directly performing dense tracking for every pixel at every frame is computationally expensive. To address this, we propose an efficient strategy for generating dense, smoothed pointmaps. First, we track a sparse set of points and smooth their 3D trajectories using adaptive weighting (Sec~\ref{sec:smooth}). Next, we propagate the displacements resulting from the smoothing process to their local neighbors through linear blending (Sec~\ref{sec:blend}), ultimately producing dense, smoothed pointmaps. 

This smoothing process is applied in a non-overlapping sliding window manner. For each local window, smoothing is performed on an extended window that includes additional frames on both ends. However, only the smoothed results within the original window are retained. This approach ensures both computational efficiency and temporal consistency.

\subsubsection{Trajectory Smoothing with Adaptive Weighting}
\label{sec:smooth}

To enhance the smoothness of 3D trajectories while mitigating the influence of outliers, we employ a 1D convolution-based smoothing process with adaptive weights. This method ensures that trajectories are refined effectively without over-smoothing salient features. The core steps of the process are described below.

\textbf{Trajectory Representation.}
The input trajectories are represented as a tensor $\mathbf{T} \in \mathbb{R}^{T \times N \times C}$, where $T$ is the number of time frames, $N$ is the number of tracked points, and $C$ is the dimensionality of the coordinates (e.g., $C = 3$ for 3D trajectories).

\textbf{Smoothing Kernel.}
A uniform smoothing kernel of size $k$ is defined as:
\begin{equation}
\mathbf{K} = \frac{1}{k} \mathbf{1}_{k},
\end{equation}
where $\mathbf{1}_{k}$ is a vector of ones with length $k$, and we set it to 5. The kernel is normalized to ensure consistent averaging across the kernel window size.

\textbf{Outlier-Aware Weighting.}
To reduce the influence of outliers, we compute a weight matrix $\mathbf{W} \in \mathbb{R}^{T \times N}$ based on the differences between consecutive trajectory points:
\begin{equation}
\Delta \mathbf{T}_{t} = \|\mathbf{T}_{t} - \mathbf{T}_{t-1}\|_2,
\end{equation}
where $\Delta \mathbf{T}_{t}$ is the norm of the difference between consecutive points. The weights are then defined as:
\begin{equation}
\mathbf{W}_{t,n} = \exp\left(-\lambda \cdot \Delta \mathbf{T}_{t,n}\right),
\end{equation}
where $\lambda$ is a smoothing factor controlling the decay of weights for larger deviations and we set it to 1. To ensure temporal alignment, the weights are padded appropriately:
\begin{equation}
\mathbf{W}_{t} = 
\begin{cases}
\mathbf{W}_{1}, & t = 1, \\
\mathbf{W}_{t-1}, & \text{otherwise}.
\end{cases}
\end{equation}

\textbf{Weighted Convolution.}
To smooth the trajectories, we apply a weighted 1D convolution to each trajectory point:
\begin{equation}
\tilde{\mathbf{T}} = \frac{\text{Conv1D}(\mathbf{T} \odot \mathbf{W}, \mathbf{K})}{\text{Conv1D}(\mathbf{W}, \mathbf{K})},
\end{equation}
where $\odot$ denotes element-wise multiplication. The convolution is applied independently for each trajectory and coordinate dimension.

The output $\tilde{\mathbf{T}}$ is a smoothed trajectory tensor with the same shape as the input, ensuring that both global and local trajectory consistency are preserved.

\subsubsection{Linear Blend Displacement (LBD) for Point Transformation}
\label{sec:blend}

After obtaining the smoothed 3D trajectories of sparse tracking points, we leverage the observation that the displacements caused by the smoothing process are approximately consistent within local regions. Inspired by linear blend skinning, we treat the smoothed tracking points as control points. To transform all other 3D points based on the displacements of these control points, we employ a Linear Blend Displacement (LBD) approach. This method calculates proximity-weighted displacements for each point by considering its $k$ nearest control points, ensuring smooth and locally influenced transformations. The detailed steps are described below.

\textbf{Problem Formulation.}
Given a set of query points $\mathbf{X} \in \mathbb{R}^{P_1 \times 3}$, control points $\mathbf{C} \in \mathbb{R}^{P_2 \times 3}$, and control displacements $\mathbf{D} \in \mathbb{R}^{P_2 \times 3}$, the goal is to compute the transformed points $\tilde{\mathbf{X}} \in \mathbb{R}^{P_1 \times 3}$ using a weighted combination of the control displacements. Here, $P_1$ is the number of query points, and $P_2$ is the number of control points.

\textbf{Nearest Neighbor Search.}
For each query point, we identify its $k$ nearest control points using the $L_2$ distance. This yields:
\begin{align}
% \mathbf{d}_{j,k} &= \|\mathbf{X}_{j} - \mathbf{C}_{k}\|^2,
\mathbf{d}_{j,k} &= \|\mathbf{X}_{j} - \mathbf{C}_{\mathbf{I}_{j,k}}\|^2, \\
\mathbf{I}_{j,k} &= \text{Indices of the } k \text{ nearest control points},
\end{align}
where $\mathbf{d}_{j,k}$ is the squared distance between the $j$-th query point and the $k$-th nearest control point. We set $k$ to 4 in our experiments.

\textbf{Weight Computation.}
We compute proximity-based weights using inverse distance weighting:
\begin{equation}
w_{j,k} = \frac{1}{\mathbf{D}_{j,k}},
\end{equation}
The weights are normalized across the $k$ nearest neighbors:
\begin{equation}
\hat{w}_{j,k} = \frac{w_{j,k}}{\sum_{k'} w_{j,k'}}.
\end{equation}

\textbf{Displacement Aggregation.}
Using the computed weights, the displacement for each query point is aggregated as linear blend of control displacements:
\begin{equation}
\mathbf{\Delta x}_{j} = \sum_{k} \hat{w}_{j,k} \mathbf{D}_{\mathbf{I}_{j,k}},
\end{equation}
where $\mathbf{d}_{j,k}$ is the displacement of the $k$-th nearest control point.

\textbf{Point Transformation.}
Finally, the transformed query points are computed by adding the aggregated displacements:
\begin{equation}
\tilde{\mathbf{X}}_{j} = \mathbf{X}_{j} + \mathbf{\Delta x}_{j}.
\end{equation}

\subsection{Implementation Details}

\paragraph{Training Details.} We train DynPT for 50,000 steps with a total batch size of 32, starting from scratch except for the 3D-aware encoder, which is initialized from DUSt3R's pretrained encoder and kept frozen during training. The learning rate is set to 5e-4, and we use the AdamW optimizer with a OneCycle learning rate scheduler~\cite{smith2019super}.

\paragraph{Inference Details.}
To ensure fast computation during motion mask calculation, we sample static points only from the latest sliding window of DynPT, as this window already includes the majority of points in the frame. The default tracking window size for DynPT is set to 16, with a stride of 4 frames. For the Point Trajectory Smoothness (PTS) objective, the default window size is 20 frames, extended by adding 5 additional frames on each end to ensure continuity and smoothness. 
And for longer videos, the window sizes for DynPT and PTS can be further extended to reduce computational costs.

\paragraph{Optimization Details.}  
The correspondence-aided optimization is performed in two stages. In the first stage, we optimize using the global alignment (GA), camera movement alignment (CMA), and camera trajectory smoothness (CTS) objectives, with respective weights $w_{\text{GA}} = 1$, $w_{\text{CMA}} = 0.01$, and $w_{\text{CTS}} = 0.01$. During this stage, the optimization targets the depth maps $\hat{\mathbf{D}}$, camera poses $\hat{\mathbf{P}}$ and camera intrinsics $\hat{\mathbf{K}}$. 
After completing the first stage, we fix the camera pose $\hat{\mathbf{P}}$ and intrinsics $\hat{\mathbf{K}}$ and proceed to optimize depth maps $\hat{\mathbf{D}}$ only in the second stage. We apply only the point trajectory smoothness (PTS) objective with weight $w_{\text{PTS}} = 1$ to further refine the depth maps $\hat{\mathbf{D}}$ in the second stage. 
Both stages are optimized for 300 iterations using the Adam optimizer with a learning rate of 0.01.

\paragraph{Datasets and Evaluation.}  
Following~\cite{hu2024depthcrafter,zhang2024monst3r}, we sample the first 90 frames with a temporal stride of 3 from the TUM-Dynamics~\cite{tum} and ScanNet~\cite{dai2017scannet} datasets for computational efficiency.
For dynamic accuracy evaluation, we use the validation sets of the MOVi-E, Panning MOVi-E, and MOVi-F datasets, comprising 250, 248, and 147 sequences, respectively. Each sequence contains 256 randomly sampled tracks spanning 24 frames. The resolution is fixed at $256 \times 256$, consistent with the TAPVid benchmark~\cite{doersch2022tap}. The evaluation metric used is accuracy, which assesses both dynamic (positive) and static (negative) states, defined as:

\begin{equation}
\text{D-ACC} = \frac{\text{TP} + \text{TN}}{\text{TP} + \text{TN} + \text{FP} + \text{FN}},
\end{equation}

where $\text{TP}$ denotes true positives, $\text{TN}$ denotes true negatives, $\text{FP}$ denotes false positives, and $\text{FN}$ denotes false negatives.

\end{document}